\documentclass[lettersize,journal]{IEEEtran}
\usepackage{amsmath,amsfonts}
\usepackage{algorithmic}
\usepackage{algorithm}
\usepackage{array}
\usepackage[caption=false,font=normalsize,labelfont=sf,textfont=sf]{subfig}
\usepackage{textcomp}
\usepackage{stfloats}
\usepackage{url}
\usepackage{verbatim}
\usepackage{graphicx}
\usepackage{cite}
\usepackage{booktabs}
\usepackage{multirow}
\usepackage{amssymb}
\usepackage{bbm}
\usepackage{bbding}
\usepackage{tcolorbox}
\usepackage{xcolor}
\usepackage{colortbl}

\definecolor{phase1color}{RGB}{240, 245, 255} 
\definecolor{phase2color}{RGB}{245, 250, 240} 

\usepackage{multirow}
\usepackage[table]{xcolor}

\newcommand{\new}[1]{\textcolor{black}{#1}}

\begin{document}

\title{ROBIN++: Unified Copyright Protection and Tamper Localization for Diffusion Models via Dual-Domain Synergistic Watermarking}



\author{Huayang Huang, Siqi Zeng, Qian Wang, Bo Du, Yu Wu


\thanks{H. Huang, S. Zeng, and B. Du are with the School of Computer Science, National Engineering Research Center for Multimedia Software, and Hubei Key Laboratory of Multimedia and Network Communication Engineering, Institute of Artificial Intelligence, Wuhan University, Wuhan, China (E-mail: hyhuang, zengsiqi, dubo@whu.edu.cn).}

\thanks{Y. Wu is with the School of Artificial Intelligence, National Engineering Research Center for Multimedia Software, and Hubei Key Laboratory of Multimedia and Network Communication Engineering, Institute of Artificial Intelligence, Wuhan University, Wuhan, China (E-mail: wuyucs@whu.edu.cn).}

\thanks{Q. Wang is with the School of Cyber Science and Engineering, Wuhan University, Wuhan, China (E-mail: qianwang@whu.edu.cn).}

\thanks{Corresponding authors: Yu Wu and Bo Du.}}



\maketitle

\begin{abstract}
Digital watermarking provides a promising solution for verifying the provenance and integrity of generated content. However, existing methods often suffer from an inherent trade-off between copyright robustness and localization sensitivity due to the entanglement of conflicting signals within a single domain.
In this paper, we propose ROBIN++, a paradigm that shifts from single-domain entanglement to a \emph{Dual-Domain Synergistic} design. Our key insight is to explicitly decouple robustness and fragility during injection by assigning them to distinct domains, while re-coupling them for mutual reinforcement during verification.
Specifically, we introduce an Adversarial Optimization Algorithm to inject a frequency-domain copyright watermark during generation, maximizing extraction robustness without visual artifacts.
Complementarily, a Content-Aware Perturbation generator is integrated into the latent space to inject the spatial-domain localization watermark, ensuring sensitivity to spatial tampering with minimal interference to the copyright watermark. 
During verification, we propose a dual-branch synergistic detector that facilitates Frequency-to-Spatial Synergy, where robust spectral priors act as global anchors for local anomaly detection. Conversely, a Spatial-to-Frequency Refinement strategy is employed to rectify corrupted latent features using predicted tamper masks.
Extensive experiments show that ROBIN++ significantly outperforms existing approaches in both copyright recovery and tamper localization accuracy, effectively mitigating the long-standing trade-off between the two tasks. Code is available at \url{https://github.com/Hannah1102/ROBIN}.
\end{abstract}


\section{Introduction}

\IEEEPARstart{T}{he} rapid evolution of diffusion models (DMs) has democratized high-fidelity content creation~\cite{ho2020denoising,saharia2022photorealistic}. However, this technological leap also creates security threats. On one front, the strong imitation capability of DMs raises serious concerns about copyright infringement~\cite{shan2023glaze,liu2024iterative}. On the other, the photorealism of generated imagery facilitates malicious tampering and disinformation (e.g., Deepfakes)~\cite{van2023anti,chen2023editshield}.

\begin{figure}[t]
  \centering
  \includegraphics[width=\linewidth]{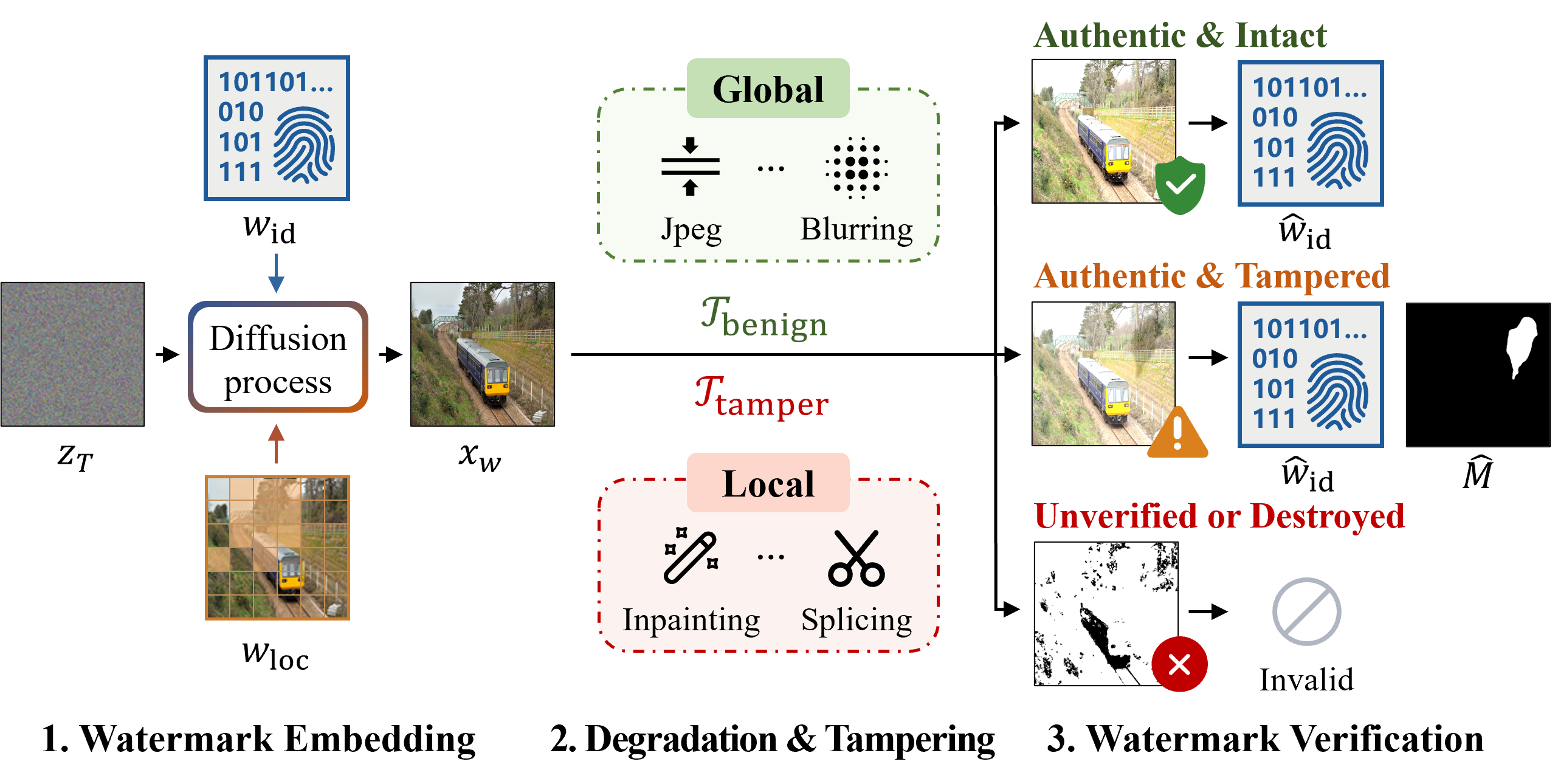}
  \caption{Overview of the unified verification framework. Dual-functional watermarking framework embeds complementary signals for copyright ($w_{\text{id}}$) and tamper localization ($w_{\text{loc}}$). Upon receiving a potentially degraded or tampered image, the system performs dual extraction to classify the image into one of three states: Authentic \& Intact, Authentic \& Tampered, or Unverified.}
  \label{fig:scenario}
  \vspace{-0.25cm}
\end{figure}

To mitigate these risks, digital watermarking~\cite{navas2008dwt,al2007combined,cheng2012refining} enables content verification by embedding imperceptible signals into generated images.
For \textit{copyright protection}, the embedded signal enables provenance verification by supporting reliable extraction of source-identifying information.
For \textit{tamper localization}, the watermark is designed to respond to local modifications, enabling identification of regions that have been maliciously altered.
In practice, these two functions are inherently coupled.
Tamper localization is meaningful only for content confirmed as one's own, while excessive tampering can undermine the reliability of copyright verification.
Motivated by this interdependence, recent efforts~\cite{editguard,zhang2024omniguard,tag} have moved towards a \textit{unified verification} setting (Fig.~\ref{fig:scenario}) to jointly ensure provenance and integrity.

\begin{figure}[t]
  \centering
  \includegraphics[width=.8\linewidth]{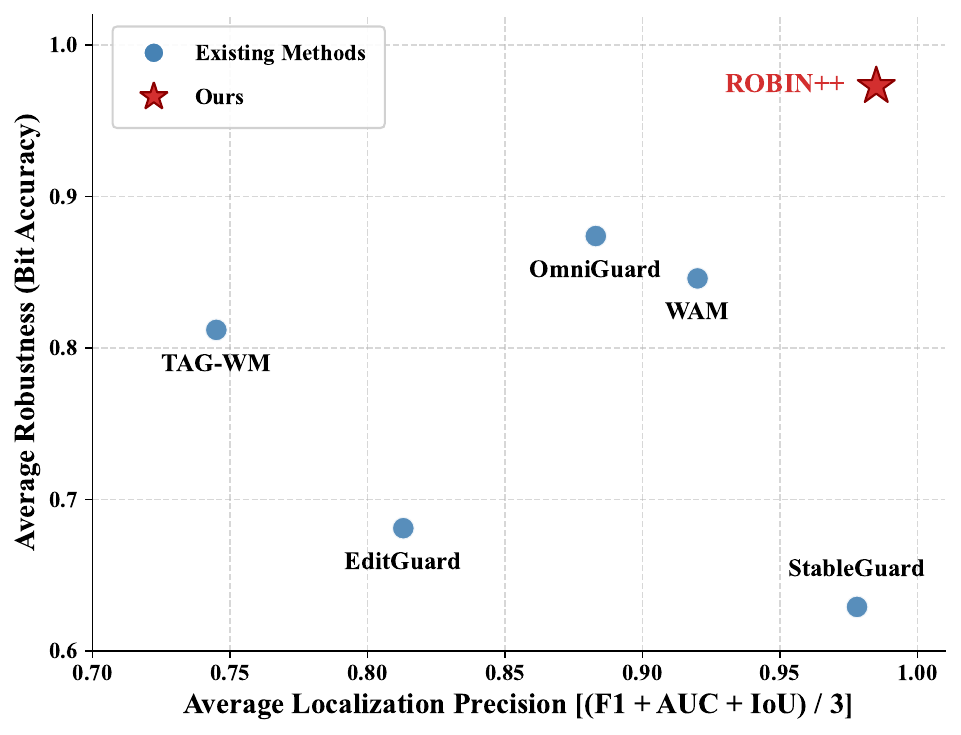}
  \caption{
  Robustness versus tamper localization precision of ROBIN++ and existing methods \cite{editguard,zhang2024omniguard,wam,tag,stableguard}.
  The results are evaluated based on the average performance across various common global degradations and local tampering operations.
  While existing methods often face a trade-off between robustness and localization, ROBIN++ achieves superior performance in both dimensions.
  }
  \label{fig:pareto}
\end{figure}

However, existing methods~\cite{editguard,zhang2024omniguard,tag,wam,stableguard} often struggle to balance copyright robustness and localization sensitivity due to the entanglement of watermark signals with contradictory requirements. 
Specifically, copyright protection requires strong robustness to withstand benign distortions, whereas tamper localization demands high sensitivity to detect subtle, localized modifications. 
Current paradigms~\cite{editguard,zhang2024omniguard,tag,stableguard,wam} attempt to address both objectives by injecting multiple watermark signals within a single domain (typically the pixel space) and decoding them independently. 
However, under such \emph{single-domain injection} designs, both robustness and sensitivity are enforced through the same signal representation and optimized over the same set of parameters.
This forces two conflicting objectives to compete within a shared optimization space, where strengthening robustness inevitably smooths or suppresses the fine-grained variations required for accurate localization, and vice versa.
Consequently, these methods exhibit an inherent performance trade-off, as illustrated in Fig.~\ref{fig:pareto}.
StableGuard~\cite{stableguard} prioritizes localization precision but suffers from poor robustness, whereas TAG-WM~\cite{tag} yields strong robustness while sacrificing the granularity required for precise localization.

\begin{figure}[t]
  \centering
  \includegraphics[width=\linewidth]{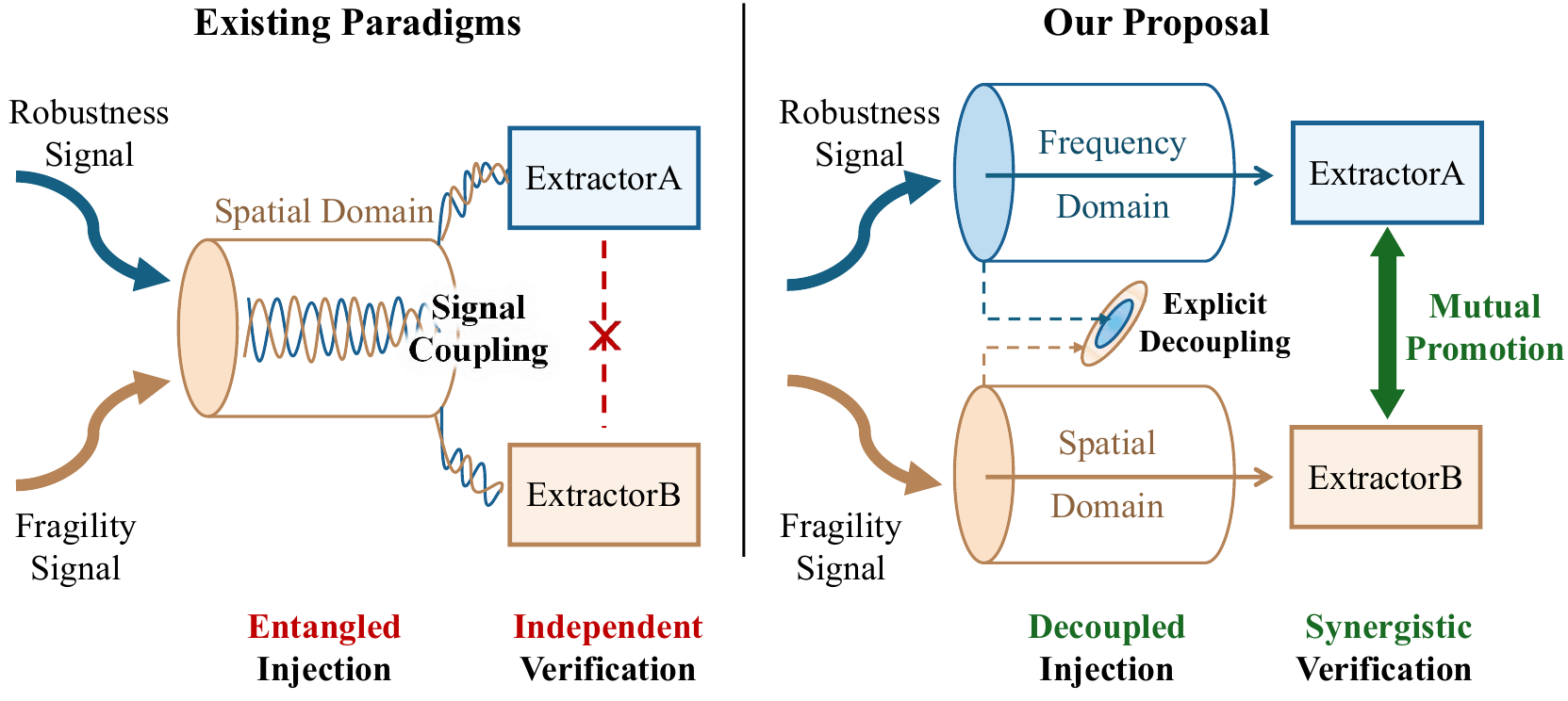}
  \caption{Conceptual comparison between existing methods and our framework.
(Left) Single-domain injection entangles robustness and localization, resulting in an inherent trade-off.
(Right) Our proposal decouples the two objectives by dual-domain injection and enables mutual reinforcement during verification.
}
  \label{fig:intro_diff}
\end{figure}

In this paper, we propose a paradigm shift from single-domain entanglement to a \textbf{Dual-Domain Synergistic} framework, which aims to mitigate this trade-off by \emph{decoupling robustness and fragility during injection and re-coupling them in a principled manner during verification}.
Unlike traditional paradigms that force conflicting requirements into a shared representation, this strategy eliminates mutual interference, allowing each watermark to be optimized within its most suitable domain without being suppressed by the other.
Specifically, as depicted in Fig.~\ref{fig:intro_diff}, we assign the robust copyright watermark to the frequency domain to exploit global stability, while embedding the fragile localization watermark in the spatial domain for precise pixel-level correspondence.
We integrate the watermarking process directly into the diffusion generation cycle and constrain the spectral distribution of both watermarks to minimize mutual interference.
During verification, a synergistic extraction strategy re-establishes their connection, enabling the two signals to provide mutual context and reinforce each other.

\textbf{Adversarial Frequency Injection for Robustness.}
For robust copyright protection, we inject a frequency-domain watermark during the intermediate stages of diffusion generation.
Existing post-hoc frequency watermarking~\cite{navas2008dwt,zhu2018hidden,cox1996secure} relies on \emph{passive} concealment after generation, which compromises watermark strength to maintain imperceptibility.
To address this, our preliminary version~\cite{huang2024robin} proposes an \emph{active} concealment strategy that exploits the iterative nature of diffusion models. Instead of superficial post-processing, we inject the watermark during intermediate generation stages and then use optimized guidance to progressively conceal the injected signal through an \emph{Adversarial Optimization Algorithm}. 
The watermark is maximized for extraction strength, while a learnable guidance is simultaneously optimized to minimize visual artifacts.
This strategy explicitly incorporates imperceptibility into the generation process, relaxing constraints on watermark strength.

\textbf{Content-Aware Spatial Injection for Localization.} 
For tamper localization, we introduce a lightweight \emph{Content-Aware Perturbation (CAP)} generator to inject a fragile spatial-domain watermark into the latent space.
Unlike post-hoc encoders~\cite{editguard, zhang2024omniguard} that modify finalized images and may interfere with existing textures, CAP operates during the generation process and produces an invisible additive perturbation conditioned on the current latent representation.
By shifting to in-process synthesis, the perturbation can adaptively manifest within optimal structural regions without the constraints of a fixed pixel space.
Furthermore, unlike encoder-based injection with unpredictable spectral distributions, our CAP is designed as an additive noise whose spectral distribution is restricted to be disjoint from that of the robust copyright watermark.
This decoupling allows both objectives to be optimized with minimal mutual interference.


\textbf{Bidirectional Synergistic Extraction.}
Beyond decoupled embedding, the extraction strategy is equally critical. However, standard approaches~\cite{editguard,zhang2024omniguard} typically extract watermarks in isolation, ignoring the mutual context between them. 
To bridge this gap, we propose a dual-branch synergistic detector that re-couples frequency- and spatial-domain cues to identify tampered regions.
The robust spectral watermark provides a global reference that helps distinguish benign global distortions (e.g., JPEG compression) from malicious local modifications.
Conversely, the predicted tamper mask is used to refine copyright verification by reducing the influence of corrupted regions during watermark extraction.
Through this bidirectional \emph{Frequency-to-Spatial Synergy} and \emph{Spatial-to-Frequency Refinement}, the two tasks can be simultaneously strengthened rather than mutually compromised.

This article extends our preliminary work ROBIN (accepted at NeurIPS 2024). While the conference version focused solely on copyright protection, this extension generalizes the framework into a unified dual-function system, hereafter referred to as ROBIN++. 
Instead of the traditional single-domain entanglement where robustness and fragility conflict, we introduce a Dual-Domain Synergistic paradigm that decouples injection for independence and re-couples extraction for mutual reinforcement.
Technically, this is implemented via the proposed Content-Aware Perturbation (CAP) for in-process spatial injection and a Synergistic Extraction architecture for cross-domain verification.
Extensive new experiments, including ablation studies on the synergy mechanism and evaluations against diverse adversarial and tampering attacks, demonstrate the systemic superiority of this extended framework.

In summary, our key contributions are as follows:
\begin{itemize}
\item  We propose a Dual-Domain Synergistic Watermarking framework that explicitly decouples robustness and fragility during injection and re-couples them during verification, mitigating the inherent conflict of traditional dual-watermarking. 
\item We introduce a lightweight Content-Aware Perturbation (CAP) generator that synthesizes controllable additive noise within the latent space, enabling adaptive localization with fewer pixel-space constraints.
\item We introduce a Synergistic Extraction strategy where copyright verification and tamper localization mutually enhance each other through a dual-branch architecture.
\item  Extensive experiments demonstrate that our approach significantly outperforms existing methods in both copyright recovery rates and tamper localization accuracy.
\end{itemize}

\section{Related Work}

\subsection{Diffusion generation and inversion}

Diffusion models~\cite{ho2020denoising,dhariwal2021diffusion,song2019generative,song2020score} generate data by iteratively reversing a progressive Gaussian diffusion process. In the latent diffusion framework, a clean latent $z_0$ is mapped to a noisy representation $z_t$ at timestep $t$ via:
\begin{equation}
z_t = \sqrt{\bar{\alpha}_t} z_0 + \sqrt{1-\bar{\alpha}_t} \epsilon, \quad \epsilon \sim \mathcal{N}(0,\mathbf{I}),
\label{eq:add_noise}
\end{equation}
where $\bar{\alpha}_t$ denotes the noise schedule. A denoising network $\epsilon_\theta$ is optimized to predict the added noise $\epsilon$ conditioned on the timestep $t$ and text embedding $\psi(p)$.

For deterministic sampling and reconstruction, Denoising Diffusion Implicit Models (DDIM)~\cite{song2020denoising} formulate the generative process as an ODE. At each step $t$, the model first estimates the denoised latent $z_0'$ and then computes the sample at $t-1$:
\begin{equation}
z_0' = \frac{z_t - \sqrt{1-\bar{\alpha}_t} \epsilon_\theta(z_t, t, \psi(p))}{\sqrt{\bar{\alpha}_t}},
\label{eq:prex0}
\end{equation}
\begin{equation}
z_{t-1} = \sqrt{\bar{\alpha}_{t-1}} z_0' + \sqrt{1-\bar{\alpha}_{t-1}} \epsilon_\theta(z_t, t, \psi(p)).
\end{equation}
The deterministic nature of DDIM trajectory allows for DDIM Inversion, which facilitates the retrieval of intermediate latent states from a generated image. This reversibility is the mathematical foundation of our robust copyright extraction, as it enables the recovery of frequency-domain watermarks embedded in the latent trajectory.





\subsection{Copyright watermarking}

The copyright watermark of generative models can be introduced either after the generation (post-processing) or during the sampling process (in-processing). Post-processing methods can adopt traditional digital image watermarking technology. 
Popular methods include frequency domain watermarking, which modifies the image representation in domains like Discrete Wavelet Transform (DWT)~\cite{xia1998wavelet} or Discrete Cosine Transform (DCT)~\cite{cox2007digital}. 
DwtDct watermarking~\cite{al2007combined} is applied in open sourced model Stable Diffusion. 
Frequency watermarks can be designed to be robust against common image manipulations like cropping, scaling, and even compression~\cite{urvoy2014perceptual}. 
However, post-processing watermarking~\cite{zhang2019robust} is constrained by limited spatial redundancy, making it difficult to achieve strong robustness without compromising imperceptibility.

In-processing methods make the watermark become part of the generated image by interfering with the generation process. 
Early approaches explored adding watermarks to training data~\cite{yu2021artificial,zhao2023recipe,cui2023diffusionshield,ditria2023hey,xiong2023flexible,fernandez2023stable}, essentially building a watermark encoder into the model. 
Recent methods simplified this process by carefully constructing the initial noise~\cite{wen2023tree,yang2024gaussian,ci2024ringid} to embed the watermark and reversing the full generation process for verification.
Compared to post-processing methods, in-processing watermarking yields superior robustness and capacity.
However, existing generative watermarking methods \cite{wen2023tree,tag} are limited in tamper localization, as initial-noise modulation introduces diffusion inversion errors.


\subsection{Tamper localization}
Tamper localization serves as an integrity verifier, requiring the precise identification of manipulated regions within a compromised image.
Existing solutions can be categorized into passive forensics and active watermarking. Passive forensics (such as MVSS-Net~\cite{mvssnet} and HDF-Net~\cite{hdfnet}) rely on detecting intrinsic anomalies, such as boundary inconsistencies and the noise view of images. However, their reliance on paired training data severely restricts their generalization capability across unseen manipulation techniques. In contrast, fragile watermarking actively injects signals to sense tampering. While early approaches~\cite{cheng2012refining,asnani2022proactive,asnani2023probed} utilized multi-scale fusion to achieve block-level localization, they often struggle to accurately identify tampering of arbitrary shapes and sizes.

Recently, several data-driven frameworks have attempted to achieve dual functionality (simultaneous copyright authentication and tamper localization).
Recent efforts like EditGuard~\cite{editguard}, OmniGuard~\cite{zhang2024omniguard} and WAM~\cite{wam} operate in a post-hoc manner, relying on training encoder-decoder networks to add watermarks after image generation.
TAG-WM~\cite{tag} and StableGuard~\cite{stableguard} inject a shared watermark for both goals during generation by modulating the initial noise and fine-tuning the decoder.
However, these approaches overlook the inherent conflict between the two watermarking objectives.
Merging these conflicting requirements into a coupled embedding space leads to mutual interference, making it difficult to optimize for both.
To bridge this gap, we disentangle these conflicting signals and inject the robust and fragile watermarks into distinct, task-specific spaces. 
This design allows for the simultaneous realization of high-performance copyright authentication and precise tamper localization without mutual interference.

\section{Methodology}

\subsection{Task Definition}
Let $x \in \mathbb{R}^{H \times W \times C}$ denote the original clean image generated by a diffusion model. Our objective is to establish a dual-functional watermarking framework that simultaneously ensures copyright authentication and tamper localization.
We introduce two distinct signals: a robust copyright watermark $w_{\text{id}}$ and a fragile localization watermark $w_{\text{loc}}$. To prevent mutual interference, these signals are injected into distinct feature spaces (e.g., frequency space and spatial space) via a composite injection function $\mathcal{I}$:
\begin{equation}
x_w = \mathcal{I}(x, w_{\text{id}}, w_{\text{loc}}),
\label{eq:implant}
\end{equation}
where $x_w$ represents the watermarked image.
During transmission, $x_w$ may undergo a compound degradation process $\mathcal{T} = \mathcal{T}_{\text{benign}} \circ \mathcal{T}_{\text{tamper}}$, which encompasses both benign global transformations $\mathcal{T}_{\text{benign}}$ (e.g., compression, resizing) and malicious local tampering $\mathcal{T}_{\text{tamper}}$ (e.g., inpainting, splicing). The received image is denoted as $\hat{x} = \mathcal{T}(x_w)$.

The model owner employs two specialized extraction mechanisms: a robust extractor $\mathcal{D}_{\text{id}}(\cdot)$ to retrieve the copyright information $\hat{w}_{\text{id}}$, and a localization detector $\mathcal{D}_{\text{loc}}(\cdot)$ to predict a tampering mask $\hat{M} \in \{0, 1\}^{H \times W}$ (where 1 indicates tampered regions). As shown in Fig.~\ref{fig:scenario}, the system logic is formalized into three distinct scenarios based on the extraction outcomes:

\textbf{Authentic \& Intact}: If $\hat{w}_{\text{id}} = w_{\text{id}}$ and $\hat{M} = \mathbf{0}$ (null mask), the image is verified as an authentic sample from the owner and has preserved its structural integrity.

\textbf{Authentic \& Tampered}: If $\hat{w}_{\text{id}} = w_{\text{id}}$ but $\hat{M} \neq \mathbf{0}$, the image is confirmed to originate from the owner but has suffered partial tampering. The mask $\hat{M}$ localizes the manipulated areas while the copyright remains verifiable.

\textbf{Unverified or Destroyed}: If $\hat{w}_{\text{id}} \neq w_{\text{id}}$ \new{(normalized distance exceeds the empirical threshold 0.5)}, the sample is either not watermarked or has undergone catastrophic degradation, rendering provenance authentication impossible.

\new{For clarity and ease of reference, the key notations used throughout our formulations are summarized in Tab. X in the Appendix.}

\subsection{Overview of ROBIN++}

The ROBIN++ watermarking framework consists of three processes: an offline optimization phase for watermark generation, an online inference phase for watermark injection, and a bidirectional synergistic extraction process for watermark verification. During the offline optimization phase, we solve for the optimal copyright watermark $w_{\text{id}}^*$ and train a content-aware perturbation generator $\mathcal{G}_{\text{loc}}$. In the subsequent online inference, these pre-optimized components are integrated into the diffusion trajectory to synthesize watermarked images. Upon receiving a query image for authentication, the verification process jointly extracts both watermarks to leverage their mutual context, enabling a robust and high-precision assessment of both identity and integrity. The overall procedure is depicted in Fig.~\ref{fig:pipeline}. 


\begin{algorithm}[t]
\caption{ROBIN++ Online Inference.}
\label{alg:inference}
\begin{algorithmic}[1]
\REQUIRE Initial noise $z_T \sim \mathcal{N}(\mathbf{0}, \mathbf{I})$, Text prompt $p$, Injection step $t_{\text{inj}}$, Optimized $(w_{\text{id}}^*, w_p^*)$, Generator $\mathcal{G}_{\text{loc}}$, VAE decoder $\text{Dec}(\cdot)$.
\ENSURE Dual-watermarked image $x_w$.
\STATE \textbf{// Stage 1: Copyright Watermark Injection}
\FOR{$t = T$ down to $1$}
    \IF{$t > t_{\text{inj}}$}
        \STATE Standard denoising: $z_{t-1} \gets \text{Step}(z_t, \epsilon_\theta(z_t, t, \psi(p)))$.
    \ELSIF{$t = t_{\text{inj}}$}
        \STATE Inject watermark: $z_t^* \gets \text{Freq-Embed}(z_t, w_{\text{id}}^*, \Omega)$.
        \STATE $z_{t-1}^* \gets \text{Step}(z_t^*, \hat{\epsilon}_\theta)$ using Eq.~\ref{eq:guided_denoise} with prompt $p$ and guidance $w_p^*$.
    \ELSE
        \STATE Guided denoising: $z_{t-1}^* \gets \text{Step}(z_t^*, \hat{\epsilon}_\theta)$ using Eq.~\ref{eq:guided_denoise}.
    \ENDIF
\ENDFOR

\STATE \textbf{// Stage 2: Localization Watermark Injection}
\STATE $z_0^* \gets z_{t=0}^*$.
\STATE $w_{\text{loc}} \gets \mathcal{G}_{\text{loc}}(z_0^*)$ \COMMENT{Generate localization watermark}.
\STATE $z_w \gets z_0^* + w_{\text{loc}}$.
\STATE $x_w \gets \text{Dec}(z_w)$.
\RETURN $x_w$.
\end{algorithmic}
\end{algorithm}

\subsubsection{Dual-Domain Injection}
We leverage the inherent iterative nature of diffusion models to embed watermarks at different stages. 
For robust watermark injection (copyright), unlike post-hoc watermarking, the optimized $w_{\text{id}}^*$ is injected into the latent trajectory at an intermediate timestep $t_{\text{inj}}$. This injection stage corresponds to the formation of detailed local structures. For a noisy latent $z_{t_\text{inj}} \in \mathbb{R}^{h \times w \times c}$, the injection is performed in the spectral domain:
\begin{equation}
z_{t_{\text{inj}}}^* = \text{Freq-Embed}(z_{t_{\text{inj}}}, w_{\text{id}}^*, \Omega),
\label{eq:freq_inject}
\end{equation}
where $\Omega$ is a frequency-domain mask. To ensure the model ``absorbs" this perturbation without visual degradation, we optimize a prompt guidance signal $w_p^*$. The subsequent denoising step $t \le t_{\text{inj}}$ is guided by both the original text prompt $\psi(p)$ and the optimized $w_p^*$:
\begin{equation}
\begin{aligned}
\hat{\epsilon}_\theta(z_t^*,t,\psi(p),w_p^*) &= \eta_1 \epsilon_\theta(z_t^*, t, \psi(p)) + \eta_2 \epsilon_\theta(z_t^*, t, w_p^*) \\
&\quad + (1-\eta_1-\eta_2)\epsilon_\theta(z_t^*, t, \emptyset),
\end{aligned}
\label{eq:guided_denoise}
\end{equation}
where $\eta_1, \eta_2$ are guidance scales. This ensures $w_{\text{id}}^*$ is seamlessly integrated into the synthesized content.


To achieve pixel-level sensitivity, the localization signal $w_{\text{loc}}$ is embedded into the fully converged latent $z_0^*$ (which already carries $w_{\text{id}}^*$ implicitly).
We employ a content-aware perturbation generator $\mathcal{G}_{\text{loc}}$ to produce the residual signal:
\begin{equation}
z_w = z_0^* + w_{\text{loc}} = z_0^* + \mathcal{G}_{\text{loc}}(z_0^*).
\label{eq:add_pert}
\end{equation}
The final watermarked image is reconstructed via the VAE decoder: $x_w = \text{Dec}(z_w)$. This placement ensures $w_{\text{loc}}$ remains fragile to spatial manipulations while not interfering with the diffusion trajectory of the robust copyright watermark.
To provide a clear execution flow of the proposed watermarking mechanism during image synthesis, we detail the step-by-step inference procedure in Alg. \ref{alg:inference}.


\subsubsection{Bidirectional Synergistic Extraction} 
The verification process is split into two processes corresponding to the extraction of the robust and fragile signals.
For copyright authentication, 
the received image $\hat{x}$ is mapped back to $\hat{z}_{t_{\text{inj}}}$ via DDIM inversion. The robust watermark $\hat{w}_{\text{id}}$ is extracted from the spectral components of the inverted latent.
To detect manipulations, $\hat{x}$ is fed into a pre-trained dual-branch synergistic detection network $\mathcal{D}_{\text{loc}}$. 
Crucially, $\mathcal{D}_{\text{loc}}$ extracts frequency cues to use the robust $w_{\text{id}}$ as a reference anchor, aiding in the isolation of local spatial anomalies $\hat{M}$.
If tampering is detected, the predicted mask $\hat{M}$ guides a refinement process that neutralizes corrupted regions using global context, thereby preventing tamper-induced false negatives in copyright verification.

\begin{figure*}[t]
  \centering
  \includegraphics[width=\linewidth]
{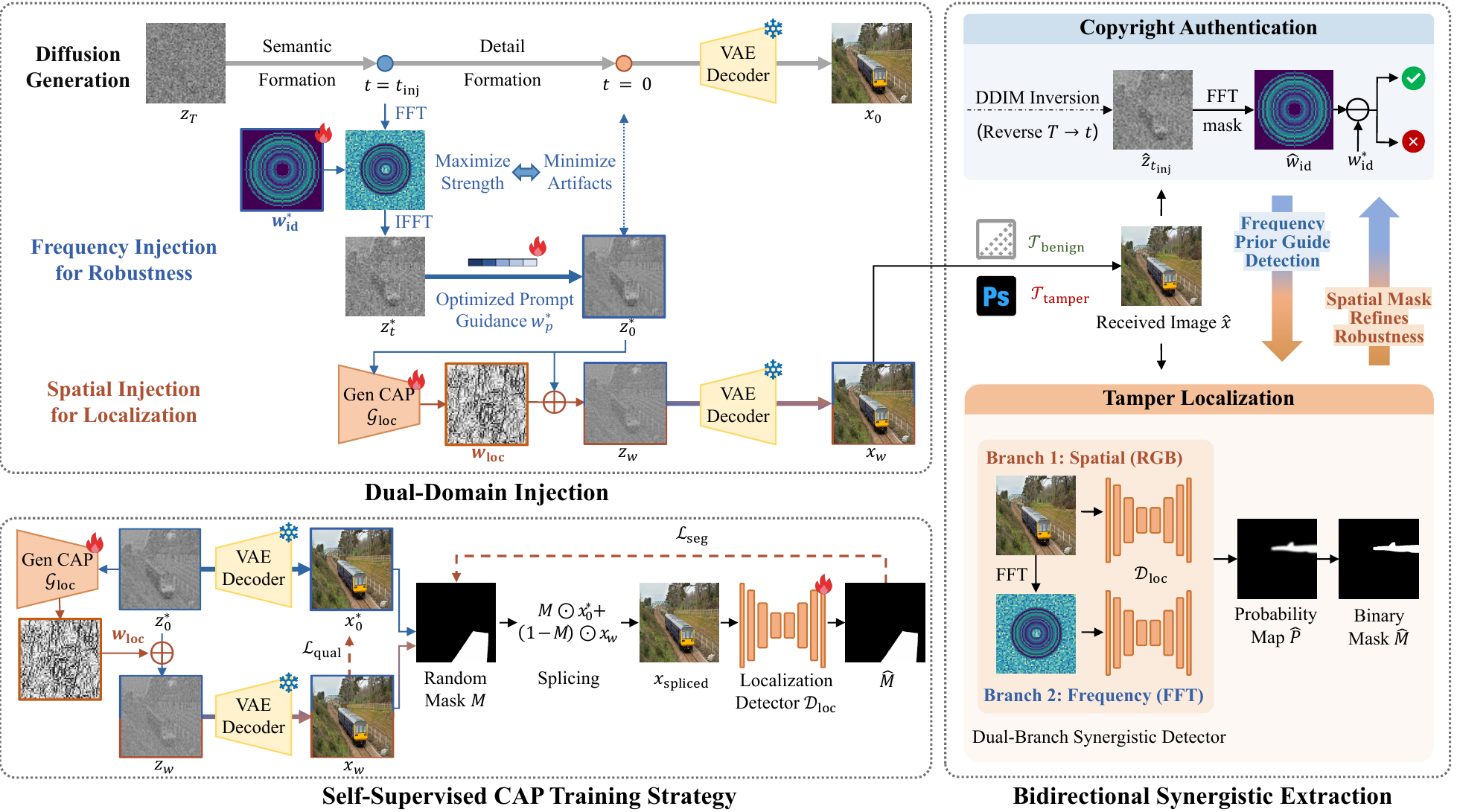}
  \caption{\textbf{The overall framework of ROBIN++.} Our approach decouples robustness and fragility through a dual-domain synergistic strategy. (Left) Dual-Domain Injection: During diffusion generation, the robust copyright watermark ($w_{\text{id}}^*$) is embedded into the frequency domain at the detail formation stage ($t_{\text{inj}} \in [200, 300]$) via Adversarial Optimization, where a learnable prompt guidance ($w_p^*$) conceals visual artifacts. Complementarily, the fragile localization signal ($w_{\text{loc}}$) is injected into the latent ($z_0^*$) at the late stage by a Content-Aware Perturbation Generator ($\mathcal{G}_{\text{loc}}$). (Bottom) Self-Supervised Training: To ensure generalizability without annotated datasets, we employ a self-supervised splicing protocol. $\mathcal{G}_{\text{loc}}$ and the Tamper Detector ($\mathcal{D}_{\text{loc}}$) are jointly optimized end-to-end by synthesizing ``hard negative'' spliced samples ($x_{\text{spliced}}$) on the fly. (Right) Bidirectional Synergistic Extraction: The watermark extraction features two interacting processes. Copyright is authenticated via diffusion inversion to retrieve frequency signals, while tampering is localized by the dual-branch detector. Crucially, the robust spectral prior aids the detector in distinguishing benign distortions from malicious tampering, while the predicted spatial mask, in turn, filters out corrupted regions to refine spectral verification, thereby achieving synergistic performance.}
  \label{fig:pipeline}
\end{figure*}

\begin{algorithm}[t]
\caption{ROBIN++ Offline Optimization.}
\label{alg:training}
\begin{algorithmic}[1]
\REQUIRE Diffusion Model $\epsilon_\theta$, VAE $\{\text{Enc}(\cdot), \text{Dec}(\cdot)\}$, Training dataset $\mathcal{X}$, Prompt set $\mathcal{P}$.
\ENSURE Optimized copyright pair $(w_{\text{id}}^*, w_p^*)$, Perturbation generator $\mathcal{G}_{\text{loc}}$, Tamper detector $\mathcal{D}_{\text{loc}}$.

\STATE \textbf{// Phase 1: Adversarial Optimization for Copyright}
\STATE Initialize $w_{\text{id}}$ randomly, $w_p \gets \psi(\text{NULL})$.
\WHILE{not converged}
    \STATE Sample $(x, p) \sim (\mathcal{X}, \mathcal{P})$ and $t \sim \mathcal{U}(200, 300)$.
    \STATE $z_t \gets \text{Forward diffusion using Eq.~\ref{eq:add_noise}}$.
    \STATE Update $w_{\text{id}}^* \gets \arg\min_{w_{\text{id}}} (\alpha \mathcal{L}_{\text{ret}} + \beta \mathcal{L}_{\text{cons}} - \lambda \| w_{\text{id}} \|)$.
    \STATE Update $w_p^* \gets \arg\min_{w_p} (\alpha \mathcal{L}_{\text{ret}} + \beta \mathcal{L}_{\text{cons}})$.
\ENDWHILE


\STATE \textbf{// Phase 2: End-to-End CAP Training for Localization}
\WHILE{not converged}
    \STATE Sample $z_0^*$ (latent containing $w_{\text{id}}^*$) from the diffusion trajectory.
    \STATE Generate $w_{\text{loc}} = \mathcal{G}_{\text{loc}}(z_0^*)$ and $z_w = z_0^* + w_{\text{loc}}$.
    \STATE \textbf{Self-Supervised Splicing}: 
    \STATE $x_w = \text{Dec}(z_w), x_{0}^* = \text{Dec}(z_0^*)$.
    \STATE Synthesize $x_{\text{spliced}} = M \odot x_0^* + (1-M) \odot x_w$.
    \STATE Compute $\mathcal{L}_{\text{loc}} = \lambda_{\text{seg}} \mathcal{L}_{\text{seg}}(\mathcal{D}_{\text{loc}}(x_{\text{spliced}}), M) + \lambda_{\text{qual}} \mathcal{L}_{\text{qual}}$.
    \STATE Update parameters of $\mathcal{G}_{\text{loc}}$ and $\mathcal{D}_{\text{loc}}$ via backpropagation.
\ENDWHILE
\RETURN $(w_{\text{id}}^*, w_p^*, \mathcal{G}_{\text{loc}}, \mathcal{D}_{\text{loc}})$
\end{algorithmic}
\end{algorithm}

\subsection{Adversarial optimization algorithm}
\label{algorithm}

We employ an adversarial optimization algorithm to jointly optimize the robust copyright watermark $w_{\text{id}}$ and the hiding prompt guidance $w_p$. The objective is twofold:  
$w_p$ is optimized to guide the model to conceal the induced visual artifacts, and $w_{\text{id}}$ tries to be as strong as possible while allowing for its targeted hiding by $w_p$. 

The objective of the prompt guiding signal is to minimize the impact of the watermark on the final generated image. 
We define the \emph{image retaining loss} $\mathcal{L}_{\text{ret}}$, which penalizes excessive deviations from the original images: 
\begin{gather}
    \mathcal{L}_{\text{ret}}=\text{MSE}(x_0'^*-x_0),\\
     x_{0}'^*=\text{Dec}(\frac{z_t^*-\sqrt{1-\bar{\alpha}_t } \epsilon _\theta (z_t^*,t,\psi(p), w_p)}{\sqrt{\bar{\alpha} _t }  }) .
\end{gather}
$x_0'^*$ is the final image predicted from the watermarked noisy latent $z_t^*$ through Eq. \ref{eq:prex0} with an additional guidance $w_p$. MSE denotes the mean squared error. 

Furthermore, as the loss incurred during DDIM inversion increases proportionally with the guidance strength~\cite{mokady2023null}, we introduce a \emph{constraint term} $\mathcal{L}_{\text{cons}}$ to prevent excessive prompt guidance introduced by $w_p$:
\begin{equation}
    \mathcal{L}_{\text{cons}} = \text{MSE}(\epsilon _\theta(z_t^*,t, w_p)-\epsilon _\theta(z_t^*,t, \psi(\emptyset))).
\end{equation}
To achieve robustness, we embed the watermark in the frequency domain of the image~\cite{wen2023tree}. Frequency domain signals are more resistant to spatial operations compared to spatial domain signals~\cite{wei2023generative}. Similar to ~\cite{wen2023tree}, we set the watermark as multiple concentric rings, but we further optimize its value to the maximum within the aforementioned constraints for greater strength and better robustness.
The spectral mask $\Omega$ in Eq.~\ref{eq:freq_inject} is strategically designed to cover the low-and-middle frequency range. This choice leverages the fact that basic image components are concentrated in these bands, providing a stable carrier for $w_{\text{id}}$ that resists noise and compression.

The optimization losses of $w_p$ and $w_\text{id}$ become
\begin{align}  
    \mathcal{L}_{w_p} &= \alpha \mathcal{L} _{\text{ret}}+\beta \mathcal{L} _{\text{cons}}, \\  
    \mathcal{L}_{w_\text{id}} &= \alpha \mathcal{L} _{\text{ret}}+\beta \mathcal{L} _{\text{cons}}-\lambda \left \| w_{\text{id}} \right \|.  
\end{align}
Since $w_\text{id}$ and $w_p$ are interdependent, we employ an alternating optimization method, in
which we iteratively optimize one while fixing the other. More
details about the optimization algorithm
are presented in Alg.~\ref{alg:training}.

\subsection{Content-aware perturbation (CAP) generator training}
\label{sec:cap}

While the robust identifier $w_{\text{id}}$ ensures provenance under global distortions, achieving precise tamper localization requires a signal that is sensitive to local spatial alterations. To this end, we introduce the Content-Aware Perturbation (CAP) mechanism, which comprises a perturbation generator $\mathcal{G}_{\text{loc}}$ and a synergistic detector $\mathcal{D}_{\text{loc}}$.
The core objective of this stage is to satisfy a dual constraint: the generated $w_{\text{loc}}$ should exhibit maximum fragility to malicious manipulations while remaining imperceptible to human observers. To balance invisibility and detectability without relying on manually annotated tampering datasets, we propose a joint training objective facilitated by a self-supervised splicing protocol.

\subsubsection{Content-Aware Perturbation Generator}
As formulated in Eq.~\ref{eq:add_pert}, the generator $\mathcal{G}_{\text{loc}}$, parameterized by a lightweight CNN~\cite{lecun2015deep}, synthesizes a residual perturbation $w_{\text{loc}}$ conditioned on the input latent $z_0^*$.
This content-awareness is crucial because it allows the generator to adaptively modulate the perturbation intensity based on local structures, thereby maximizing the localization sensitivity without compromising the generative quality.
\new{The task of synthesizing manipulation-sensitive perturbations primarily relies on capturing local fine-grained spatial patterns, making a lightweight CNN highly sufficient.}
To prevent $w_{\text{loc}}$ from interfering with the mid-to-low frequency copyright information, we apply a high-pass filter to $w_{\text{loc}}$, ensuring that its energy is primarily concentrated in the high-frequency components.
To ensure the perturbation remains imperceptible, we enforce a \emph{quality constraint loss} $\mathcal{L}_{\text{qual}}$ between the decoded images of the original and perturbed latents:
\begin{equation}
    \mathcal{L}_{\text{qual}} = \| \text{Dec}(z_{w}) - \text{Dec}(z_0^*) \|_2^2 + \gamma \cdot \text{LPIPS}(\text{Dec}(z_{w}), \text{Dec}(z_0^*)),
    \label{eq:l_qual}
\end{equation}
where $\text{Dec}(\cdot)$ is the VAE decoder and $\gamma$ balances pixel-wise and perceptual fidelity.

\subsubsection{Self-Supervised Splicing Protocol}
A major bottleneck in existing methods is the difficulty of constructing large-scale, diverse tampered datasets. We address this by proposing a self-supervised splicing strategy that leverages the subtle variance between the watermarked image $x_{w} = \text{Dec}(z_{w})$ and its non-watermarked counterpart $x_{0}^* = \text{Dec}(z_0^*)$.
We synthesize a tampered sample $x_{\text{spliced}}$ by applying a random binary mask $M \in \{0, 1\}^{H \times W}$:
\begin{equation}
    x_{\text{spliced}} = M \odot x_0^* + (1-M) \odot x_{w}.
\end{equation}
Here, $M$ indicates the tampered region where the fragile localization watermark is removed or replaced by the non-localization-watermarked counterpart, while the $(1-M)$ region remains watermarked.
Since $x_{w}$ and $x_{0}^*$ are visually nearly identical, splicing them creates ``hard negative" samples where the tampering boundaries are extremely subtle. This forces the detector to learn intrinsic watermark fragility rather than relying on obvious visual artifacts, thereby improving generalization to unseen attacks.

\subsubsection{Dual-Branch Synergistic Detector}
The detector $\mathcal{D}_{\text{loc}}$ aims to predict the tampering mask $\hat{M}$ given $x_{\text{spliced}}$. We employ a U-Net architecture~\cite{ronneberger2015u} with a dual-branch encoder to capture anomalies in both spatial and frequency domains.
The spatial branch processes RGB pixels to capture local discontinuities. The frequency branch takes the FFT-based spectral features of the input, aiming to achieve \emph{frequency-to-spatial synergy} by leveraging the underlying robust watermark $w_\text{id}^*$ as a reference anchor. Since $w_\text{id}^*$ is frequency-based and robust, consistency checks in the spectral domain can aid in distinguishing between benign global degradation and malicious local tampering.
The features from both branches are fused via concatenation and passed to the decoder to predict the tampering probability $\hat{P} = \mathcal{D}_{\text{loc}}(x_{\text{spliced}})$, from which tampering mask $\hat{M}$ is derived.

\subsubsection{Optimization Objectives}
To address the class imbalance between tampered regions and the background, we adopt a hybrid segmentation loss $\mathcal{L}_{\text{seg}}$ combining pixel-wise Binary Cross-Entropy ($\mathcal{L}_{\text{bce}}$) and Dice loss ($\mathcal{L}_{\text{dice}}$):
\begin{equation}
    \mathcal{L}_{\text{seg}}(\hat{M}, M) = \lambda_{\text{bce}} \mathcal{L}_{\text{bce}}(\hat{M}, M) + \lambda_{\text{dice}} \mathcal{L}_{\text{dice}}(\hat{M}, M),
    \label{eq:l_seg}
\end{equation}
where $\lambda_{\text{bce}}$ and $\lambda_{\text{dice}}$ are balancing weights.
Specifically, the Binary Cross-Entropy loss ensures pixel-level classification accuracy and is defined as:
\begin{equation}
\begin{aligned}
\mathcal{L}_{\text{bce}}(\hat{M}, M)
= - \frac{1}{H W}
\sum_{i=1}^{H} \sum_{j=1}^{W}
\Big[
&\, M_{i,j} \log(\hat{M}_{i,j})
\\
&+ (1 - M_{i,j})
\log(1 - \hat{M}_{i,j})
\Big].
\end{aligned}
\end{equation}
where $M_{i,j}$ and $\hat{M}_{i,j}$ denote the ground truth and predicted tampering masks at coordinate $(i, j)$.
To further enforce structural consistency and mitigate the bias towards the dominant background class, we incorporate the Dice loss:
\begin{equation}
    \mathcal{L}_{\text{dice}}(\hat{M}, M) = 1 - \frac{2 \sum_{i,j} (M_{i,j} \cdot \hat{M}_{i,j}) + \epsilon}{\sum_{i,j} M_{i,j} + \sum_{i,j} \hat{M}_{i,j} + \epsilon},
\end{equation}
where $\epsilon$ is a smooth term (set to $1e-5$) to prevent division by zero and improve numerical stability during training. By jointly optimizing these objectives, the detector is encouraged to produce sharp boundaries while maintaining high recall on small tampered regions.

The framework is trained end-to-end. 
The total joint training objective for this stage is:
\begin{equation}
    \mathcal{L}_{\text{loc}} = \lambda_{\text{seg}} \mathcal{L}_{\text{seg}}(\hat{M}, M) + \lambda_{\text{qual}} \mathcal{L}_{\text{qual}}.
    \label{eq:l_total}
\end{equation}
This joint optimization ensures the perturbation is both invisible to the human eye and highly discriminative for the detector.
The complete optimization procedure, encompassing both the adversarial optimization and the self-supervised localization training, is summarized in Alg.\ref{alg:training}.



\subsection{Finding keypoints for implantation}

\subsubsection{Injection of Copyright Watermark}
The selection of the injection timestep $t_{\text{inj}}$ for the copyright watermark is pivotal for balancing watermark robustness and semantic fidelity. Prior studies~\cite{yang2023diffusion,kwon2022diffusion} indicate that early diffusion stages ($t \to T$) establish global structures, while late stages ($t \to 0$) refine local details.
We perform watermark insertion within an intermediate details formation window $[200, 300]$. This choice is motivated by the observation~\cite{shallowdiffuse} that timesteps around $0.2T$-$0.3T$ exhibit low-rank Jacobian properties, where perturbations can reside in the null space of the diffusion manifold. At this stage, the semantic layout has stabilized, yet the model retains sufficient denoising flexibility to absorb frequency-domain perturbations without incurring visible artifacts.
\new{An adaptive alternative to this fixed injection window is further studied in Appendix F.}

\subsubsection{Injection of Localization Watermark}

Unlike the trajectory-embedded copyright identifier, the localization signal $w_{\text{loc}}$ requires strict spatial alignment and high sensitivity to pixel-level tampering. Injecting $w_{\text{loc}}$ during the diffusion trajectory would subject it to stochastic denoising updates, potentially destroying the pixel-to-latent correspondence.
Consequently, we target the post-denoising stage at $z_0^*$. By implanting the signal into the fully converged latent space, we leverage the VAE decoder as a non-linear upsampler to propagate the perturbation to local pixel neighborhoods. This different placement ensures that the localization signal is highly fragile to manipulations while avoiding interference with the robust watermark's trajectory.

\subsection{Watermark validation}
As illustrated in the right panel of Fig.~\ref{fig:pipeline}, the verification process features two interacting processes: a rigorous diffusion inversion process for copyright authentication and a rapid, feed-forward detection process for tamper localization.

\subsubsection{Copyright Authentication via Inversion}
To verify ownership, we perform DDIM inversion on the received image $\hat{x}$ to retrieve the latent state $\hat{z}_{t_{\text{inj}}}$ at the injection step. Since the original prompt is typically unavailable during blind verification, we follow~\cite{wen2023tree} by employing a null-text embedding with a guidance scale of 1.0. The watermark $\hat{w}_{\text{id}}$ is extracted from the spectral components of $\hat{z}_{t_{\text{inj}}}$ within the masked region $\Omega$. Authentication is confirmed if the $\ell_1$ distance between the extracted $\hat{w}_{\text{id}}$ and the reference $w_{\text{id}}$ falls below a threshold $\tau_i$:
\begin{equation}
D(w_{\text{id}}, \hat{w}_{\text{id}}) = \frac{1}{|\Omega|} \sum_{k \in \Omega} | w_{\text{id}}^k - \hat{w}_{\text{id}}^k | \le \tau_i,
\end{equation}
where $k$ denotes the spectral index within $\Omega$. This inversion-based extraction is compatible with any reversible ODE solvers such as DPM-Solver~\cite{lu2022dpm}, DPM-Solver++~\cite{lu2022dpm++}, PNDM~\cite{liupseudo}, and AMED-Solver~\cite{zhou2024fast}.


\subsubsection{Tamper Localization}
\label{sec:verify_loc}
The localization signal is detected directly in the pixel domain. Given $\hat{x}$, the pre-trained detector $\mathcal{D}_{\text{loc}}$ predicts a tampering probability map $\hat{P} = \mathcal{D}_{\text{loc}}(\hat{x})$. 
$\hat{P} \in [0, 1]^{H \times W}$ is the predicted probability map indicating the likelihood of tampering for each pixel.
The final binary mask $\hat{M}$ is obtained via thresholding: $\hat{M} = \mathbbm{1}(\hat{P} > \tau_{\text{loc}})$. Unlike passive forensics, our detector leverages the disruption of the fragile perturbation $w_{\text{loc}}$. By utilizing the frequency components of the robust identifier $w_{\text{id}}^*$ as a reference anchor, $\mathcal{D}_{\text{loc}}$ can effectively distinguish global benign distortions from local malicious manipulations, ensuring precise tamper localization.



\subsubsection{Spatial-to-Frequency Refinement}To mitigate the impact of localized tampering on copyright verification, we introduce a mask-guided spectral refinement mechanism. 
Since the robust copyright watermark $w_{\text{id}}^*$ is embedded within the latent space, the refinement process is strategically performed at the latent level.
Specifically, we derive a latent-space mask $\hat{M}_z$ by downsampling pixel-space mask $\hat{M}$ and construct a rectified latent $\tilde{z}_0$ by replacing corrupted regions with the global statistics of the intact areas:\begin{equation}\tilde{z}_{0}^{(i,j)} =
\begin{cases}
\hat{z}_{0}^{(i,j)}, & \text{if } \hat{M}_z^{(i,j)} = 0, \\
\bar{z}_{\text{intact}}, & \text{if } (\hat{M}_z)^{(i,j)} = 1,
\end{cases}
\label{eq:z0_refine}
\end{equation}
where $\bar{z}_{\text{intact}}$ denotes the spatial average of non-tampered latent features. This statistical filling prevents the introduction of artificial high-frequency edges-common in zero-padding-that would otherwise contaminate the spectral watermark. This bidirectional interaction completes the system's synergy: while spectral priors facilitate spatial localization, the localized spatial predictions, in turn, suppress adversarial spectral contamination. This loop significantly improves authentication reliability under mixed benign and malicious degradations.

\section{Experiments}

\subsection{Experimental setting}
\label{sec:setting}
\noindent \textbf{Model and dataset.}
We conduct experiments on latent diffusion models Stable Diffusion-v2~\cite{sd}. 
We optimize the adversarial optimization framework using the stable-diffusion-prompts dataset from Gustavosta~\cite{GustavostaStableDiffusionPromptsDatasets2023}. For Content-Aware Perturbation (CAP) generator training, we employ the MS-COCO training set~\cite{coco} in a fully self-supervised manner, which requires no manual supervision.

\noindent \textbf{Evaluation metrics.}
\new{\textbf{\emph{1) For robustness evaluation}}, we report two metrics: Bit Accuracy and AUC. For multi-bit baselines, `Bit Accuracy' refers to the payload decoding rate, which also serves as the decision statistic for calculating AUC. For zero-bit methods including ROBIN++, `Bit Accuracy' denotes the binary detection accuracy (watermarked vs. clean), while the AUC is derived from the L1 distance of the frequency pattern.
Evaluating both paradigms under this unified AUC framework ensures a fair assessment of the separability of watermarked and clean images.}
Specifically, we compute metrics using 1,000 watermarked and 1,000 clean images.
\textbf{\emph{2) For tamper localization precision}}, we employ three pixel-level metrics: F1 score, AUC, and IoU~\cite{mvssnet,guillaro2023trufor}. 
We conduct evaluations on 5,000 image-mask pairs from the StableGuard dataset~\cite{stableguard}. 
The tampered regions contain content synthesized by five inpainting models: Stable Diffusion~\cite{sd}, SDXL~\cite{sdxl}, Kandinsky~\cite{kandinsky}, ControlNet~\cite{controlnet}, and LaMa~\cite{lama}.
\textbf{\emph{3) For image quality evaluation}}, we employ PickScore~\cite{kirstain2023pick} to evaluate both the fidelity of the watermarked image distribution and the alignment between generated images and their corresponding text prompts. 

\begin{table*}[t!]
    \centering
    \caption{Comparison of robustness against image degradations (Bit Acc/AUC), Inpainting Localization (average of F1, AUC, and IoU), and image quality (PickScore) on generative-model-based inpainting datasets~\cite{sd, sdxl, kandinsky, controlnet, lama}. The best and second-best results are marked in \textbf{bold} and \underline{underlined}, respectively.}
    \renewcommand{\arraystretch}{1.2}
    \resizebox{\linewidth}{!}{
        \begin{tabular}{@{}l|c|ccccc|cccccc|c@{}}
            \hline
            \multirow{2}{*}{Method} & \multirow{2}{*}{Reference}
            & \multicolumn{5}{c|}{\textbf{Robustness (Bit Acc~$\uparrow$ / AUC~$\uparrow$)}} 
            & \multicolumn{6}{c|}
            {\textbf{Inpainting Localization [(F1~$\uparrow$ + AUC~$\uparrow$ + IoU~$\uparrow$) / 3]}}
            & \textbf{Quality} \\ 
          
            \cline{3-14} 
            
            & & Clean 
            & Rotation
            & JPEG 
            & Blur 
            & \textbf{Avg} 
            & SD Inpaint
            & SDXL
            & Kandinsky
            & ControlNet
            & LaMa
            & \textbf{Avg} 
            & PickScore \\ 
            \hline

            MVSS-Net~\cite{mvssnet} & TPAMI'22
            & - & - & - & - 
            & - 
            & 0.862 & 0.851 & 0.850 & 0.856 & 0.861 
            & 0.856 
            & - \\ 

            HDF-Net~\cite{hdfnet} & TPAMI'24
            & - & - & - & - 
            & - 
            & 0.595 & 0.598 & 0.587 & 0.593 & 0.603 
            & 0.595 
            & - \\ 

            EditGuard~\cite{editguard} & CVPR'24
            & \underline{0.999} / \textbf{1.000} & 0.499 / 0.474 & 0.746 / \underline{0.994} & 0.480 / 0.414 & 0.681 / 0.721 
            & 0.833 & 0.803 & 0.808 & 0.819 & 0.804 
            & 0.813 
            & \underline{19.319} \\ 

            OmniGuard \cite{zhang2024omniguard} & CVPR'25
            & \textbf{1.000} / \textbf{1.000} & 0.502 / \underline{0.535} & \underline{0.997} / \textbf{1.000} & \underline{0.998} / \textbf{1.000} & \underline{0.874} / \underline{0.884} 
            & 0.876 & 0.888 & 0.888 & 0.879 & 0.885 
            & 0.883 
            & \textbf{19.354} \\ 

            WAM \cite{wam} & ICLR'25
            & \textbf{1.000} / \textbf{1.000} & 0.481 / 0.421 & 0.904 / 0.990 & \underline{0.998} / \textbf{1.000} & 0.846 / 0.853 
            & 0.923 & 0.919 & 0.921 & 0.918 & 0.918 
            & 0.920 
            & 19.258 \\ 

            TAG-WM \cite{tag} & ICCV'25
            & \underline{0.999} / \textbf{1.000} & 0.500 / 0.516 & 0.957 / \textbf{1.000} & 0.791 / \underline{0.962} & 0.812 / 0.870 
            & 0.761 & 0.729 & 0.713 & 0.766 & 0.753 
            & 0.744 
            & 19.314 \\ 
            
            StableGuard \cite{stableguard} & NeurIPS'25
            & \textbf{1.000} / \textbf{1.000} & \underline{0.507} / 0.493 & 0.499 / 0.494 & 0.510 / 0.492 & 0.629 / 0.620 
            & \underline{0.978} & \underline{0.978} & \underline{0.977} & \underline{0.979} & \underline{0.978} 
            & \underline{0.978} 
            & 19.086 \\ 

            \hline

            ROBIN++ & -
            & \textbf{1.000} / \textbf{1.000} & \textbf{0.892} / \textbf{0.935} & \textbf{1.000} / \textbf{1.000} & \textbf{1.000} / \textbf{1.000} & \textbf{0.973} / \textbf{0.984} 
            & \textbf{0.988} & \textbf{0.984} & \textbf{0.981} & \textbf{0.988} & \textbf{0.986} 
            & \textbf{0.985} 
            & 19.124 \\ 
            \hline
      
        \end{tabular}
    }
    \label{tab:combined_comparison}
\end{table*}

\begin{table*}[t]
\centering
\caption{
Comprehensive evaluation of ROBIN++ and state-of-the-art baselines under diverse image degradations and security attacks.
We report Copyright Robustness (Bit Acc.~ / AUC, higher is better) for Imperceptible, Frequency-domain, Advanced GenAI Removal, and Perturbation-removal attacks, and Forgery Success Rate (FSR, lower is better) for Black-box Forgery attacks.
The best and second-best results are marked in \textbf{bold} and \underline{underlined}, respectively.
}
\renewcommand{\arraystretch}{1.2}
\resizebox{\linewidth}{!}{
\setlength{\tabcolsep}{4pt}
\begin{tabular}{@{}l|c|ccc|cccc|c|c@{}}
\hline
\multirow{2}{*}{\textbf{Method}}
& \textbf{Imperceptible} $\uparrow$
& \multicolumn{3}{c|}{\textbf{Frequency-domain Attacks} $\uparrow$}
& \multicolumn{4}{c|}{\textbf{Generative-AI-based Removal} $\uparrow$}
& \textbf{Perturbation-removal} $\uparrow$
& \textbf{Black-box Forgery} $\downarrow$\\
\cline{2-11}
& Adv. Noise~\cite{an2024waves}
& Freq. Noise & Low Pass & Freq. Smooth
& Latent Remv.~\cite{muller2025black} & Diffusion Rec.~\cite{zhao2024invisible} & VAE Rec.~\cite{zhao2024invisible} & Rinse2x~\cite{zhao2024invisible} & Pert. Defeat~\cite{foerster2025lightshed}
& Latent Forg.~\cite{muller2025black} \\
\hline
EditGuard~\cite{editguard}
& 0.998 / \textbf{1.000}
& \underline{0.999} / \textbf{1.000} & 0.501 / 0.483 & 0.997 / \textbf{1.000}
& 0.505 / 0.501 & 0.488 / 0.442 & 0.501 / 0.500 & 0.495 / 0.463 & \textbf{1.000} / \textbf{1.000} &
\underline{0.458} \\
OmniGuard~\cite{zhang2024omniguard}
& \underline{0.999} / \textbf{1.000}
& \textbf{1.000} / \textbf{1.000} & \textbf{1.000} / \textbf{1.000} & 0.918 / \textbf{1.000}
& 0.507 / 0.542 & 0.871 / \textbf{1.000} & \underline{0.998} / \textbf{1.000} & 0.750 / \textbf{1.000} & \textbf{1.000} / \textbf{1.000} &
0.991 \\

WAM~\cite{wam}
& 0.998 / \textbf{1.000}
& \underline{0.999} / \textbf{1.000} & \underline{0.999} / \textbf{1.000} & 0.993 / \textbf{1.000}
& 0.496 / 0.500 & 0.716 / \underline{0.857} & 0.705 / \underline{0.922} & 0.639 / 0.780 & \underline{0.999} / \textbf{1.000} & 
0.500 \\

TAG-WM~\cite{tag}
& 0.973 / \textbf{1.000}
& \underline{0.999} / \textbf{1.000} & 0.969 / \textbf{1.000} & 0.978 / \textbf{1.000}
& \underline{0.820} / \underline{0.981} & \underline{0.973} / \textbf{1.000} & 0.953 / \textbf{1.000} & \underline{0.936} / \textbf{1.000} & \underline{0.999} / \textbf{1.000} &
0.614 \\

StableGuard~\cite{stableguard}
& \textbf{1.000} / \textbf{1.000}
& \textbf{1.000} / \textbf{1.000} & 0.501 / \underline{0.494} & \underline{0.999} / \textbf{1.000}
& 0.503 / 0.508 & 0.502 / 0.500 & 0.498 / 0.488 & 0.501 / 0.504 & \textbf{1.000} / \textbf{1.000} &
0.504 \\
\hline
ROBIN++
& \textbf{1.000} / \textbf{1.000}
& \textbf{1.000} / \textbf{1.000} & \textbf{1.000} / \textbf{1.000} & \textbf{1.000} / \textbf{1.000}
& \textbf{0.975} / \textbf{0.995} & \textbf{1.000} / \textbf{1.000} & \textbf{1.000} / \textbf{1.000} & \textbf{0.968} / \underline{0.993} & \textbf{1.000} / \textbf{1.000} &
\textbf{0.411} \\
\hline
\end{tabular}
}
\label{tab:more_robust}
\end{table*}

\noindent \textbf{Implementation details.}
We employ the DPM-Solver scheduler~\cite{lu2022dpm} with a guidance scale of 7.5 and inference steps of 50. We optimize the frequency-domain watermark and the hiding prompt guidance using 50 generated images. The learning rates for the frequency watermark and prompt guidance are 0.8 and 5e-04, respectively, with a total of 1,000 optimization rounds. The default frequency watermark covers 70\% of the image frequency domain. 
For the CAP framework, we perform end-to-end training with images resized to $512 \times 512$. Optimization employs AdamW~\cite{loshchilovdecoupled} (weight decay $10^{-2}$) with learning rates of $10^{-3}$ for the generator and $10^{-4}$ for the detector. 
We configure the loss hyperparameters (Sec. 3.3) as follows: the quality weight is $\gamma = 1$ (Eq.~\ref{eq:l_qual}), and segmentation weights are $\lambda_{\text{bce}} = 0.2, \lambda_{\text{dice}} = 0.8$ (Eq.~\ref{eq:l_seg}). The overall objective weights are balanced at $\lambda_{\text{seg}} = \lambda_{\text{qual}} = 1$ (Eq.~\ref{eq:l_total}).
To prevent quality degradation, the perturbation intensity is bounded by $\xi = 0.5$.
The localization threshold $\tau_{\text{loc}}$ is set to $0.5$.
All experiments are conducted on the NVIDIA GeForce RTX 4090 GPU.

\subsection{Robustness comparison}
We compare our approach with five representative methods that support \emph{dual-function} watermarking, including EditGuard~\cite{editguard}, OmniGuard~\cite{zhang2024omniguard}, WAM~\cite{wam}, TAG-WM~\cite{tag}, and StableGuard~\cite{stableguard}.
To ensure the watermark's resilience in real-world scenarios, we delve into its robustness under various image transformations. 
These include random rotation of 75 degrees, JPEG compression with quality 25, and Gaussian blur with radius 5.
These settings are strict for watermark verification because the image has been significantly altered. 

\begin{table*}[t]
    \centering
    \caption{Localization precision comparison on the generative-model-based tampering dataset. The best and second-best results are highlighted in \textbf{bold} and \underline{underlined}, respectively.}
    \renewcommand{\arraystretch}{1.2}
    \resizebox{\linewidth}{!}{
        \begin{tabular}{l|c|ccc|ccc|ccc|ccc|ccc}
            \hline
            
            \multirow{2}{*}{Method} & \multirow{2}{*}{Reference}
            & \multicolumn{3}{c|}{SD Inpaint \cite{sd}}
            & \multicolumn{3}{c|}{SDXL \cite{sdxl}}
            & \multicolumn{3}{c|}{Kandinsky \cite{kandinsky}}
            & \multicolumn{3}{c|}{ControlNet \cite{controlnet}}
            & \multicolumn{3}{c}{LaMa \cite{lama}} \\ 
            
            \cline{3-17} 
        
            & & F1~$\uparrow$ & AUC~$\uparrow$ & IoU~$\uparrow$
            & F1~$\uparrow$ & AUC~$\uparrow$ & IoU~$\uparrow$
            & F1~$\uparrow$ & AUC~$\uparrow$ & IoU~$\uparrow$
            & F1~$\uparrow$ & AUC~$\uparrow$ & IoU~$\uparrow$
            & F1~$\uparrow$ & AUC~$\uparrow$ & IoU~$\uparrow$ \\
   
            \hline
            
            MVSS-Net \cite{mvssnet} & TPAMI'22
            & 0.862 & 0.934 & 0.791
            & 0.848 & 0.929 & 0.775
            & 0.848 & 0.928 & 0.775
            & 0.856 & 0.930 & 0.782
            & 0.860 & 0.934 & 0.789 \\
            
            HDF-Net \cite{hdfnet} & TPAMI'24
            & 0.556 & 0.762 & 0.468
            & 0.763 & 0.470 & 0.560 
            & 0.544 & 0.759 & 0.457
            & 0.551 & 0.764 & 0.463
            & 0.565 & 0.767 & 0.476 \\
            
            EditGuard \cite{editguard} & CVPR'24
            & 0.806 & 0.985 & 0.709 & 0.766 & 0.981 & 0.662 & 0.772 & 0.983 & 0.669 & 0.787 & 0.983 & 0.687 & 0.767 & 0.978 & 0.666 \\
            
            OmniGuard \cite{zhang2024omniguard} & CVPR'25
            & 0.853 & 0.964 & 0.810
            & 0.867 & 0.973 & 0.824
            & 0.868 & 0.966 & 0.830
            & 0.858 & 0.965 & 0.815
            & 0.864 & 0.969 & 0.823 \\
            
            WAM \cite{wam} & ICLR'25
            & 0.924 & 0.977 & 0.868
            & 0.918 & 0.976 & 0.862
            & 0.921 & 0.976 & 0.865
            & 0.917 & 0.977 & 0.860
            & 0.922 & 0.967 & 0.864 \\ 

            TAG-WM \cite{tag} & ICCV'25
            & 0.765 & 0.859 & 0.659 & 0.729 & 0.840 & 0.618 & 0.712 & 0.829 & 0.597 & 0.771 & 0.863 & 0.665 & 0.757 & 0.857 & 0.646 \\
            
            StableGuard \cite{stableguard} & NeurIPS'25
            & \underline{0.980} & \underline{0.993} & \underline{0.962}
            & \underline{0.981} & \underline{0.991} & \underline{0.961}
            & \underline{0.980} & \underline{0.992} & \underline{0.960}
            & \underline{0.981} & \underline{0.993} & \underline{0.963}
            & \underline{0.979} & \underline{0.993} & \underline{0.961} \\ 

            \hline

            ROBIN++ & -
            & \textbf{0.988} & \textbf{1.000} & \textbf{0.976}
            & \textbf{0.984} & \textbf{1.000} & \textbf{0.969}
            & \textbf{0.981} & \textbf{1.000} & \textbf{0.963}
            & \textbf{0.988} & \textbf{1.000} & \textbf{0.976}
            & \textbf{0.986} & \textbf{1.000} & \textbf{0.973} \\
            
            \hline
        \end{tabular}
    }
    \label{tab:aigc_dst}
    \vspace{-10pt}
\end{table*}

The comparison of copyright watermark robustness is summarized in Tab.~\ref{tab:combined_comparison}. ROBIN++ consistently achieves the highest verification accuracy across most degradations, yielding a state-of-the-art average AUC of 0.984. Notably, under image rotation, while baselines drop to near-random levels ($\sim 0.50$), our method maintains a high AUC of 0.935 by exploiting rotation-invariant spectral patterns.
\new{To further validate the robustness of ROBIN++ against advanced threats, we comprehensively evaluate its performance under imperceptible adversarial perturbations~\cite{an2024waves}, frequency-domain, GenAI-based removal~\cite{zhao2024invisible}, perturbation-removal~\cite{foerster2025lightshed}, and black-box forgery attacks~\cite{muller2025black}. As shown in Tab.~\ref{tab:more_robust}, ROBIN++ maintains competitive or superior robustness across all evaluated attacks, demonstrating clear advantages over baseline methods. 
This confirms that embedding the watermark within the intermediate generation trajectory structurally binds it to the image content, making it highly resilient to attacks. }
Furthermore, by decoupling the copyright and localization watermarks in both the injection phase and domain, we substantially reduce mutual interference between the two signals.

\subsection{Localization comparison}
In addition to the above baselines, to further evaluate the effectiveness of tamper localization, we also compare against two \emph{passive} tamper detection methods MVSS-Net~\cite{mvssnet} and HDF-Net~\cite{hdfnet} that do not rely on watermark injection.
We primarily focus on the challenging inpainting-based tampering~\cite{saharia2022palette} constructed by generative models, where the tampered regions are synthesized via conditional generation. The inpainting area is randomly sampled to cover approximately 1\% to 60\% of the image, enabling a controlled yet diverse range of tampering extents.

\begin{figure*}[t]
  \centering
  \includegraphics[width=1\linewidth]{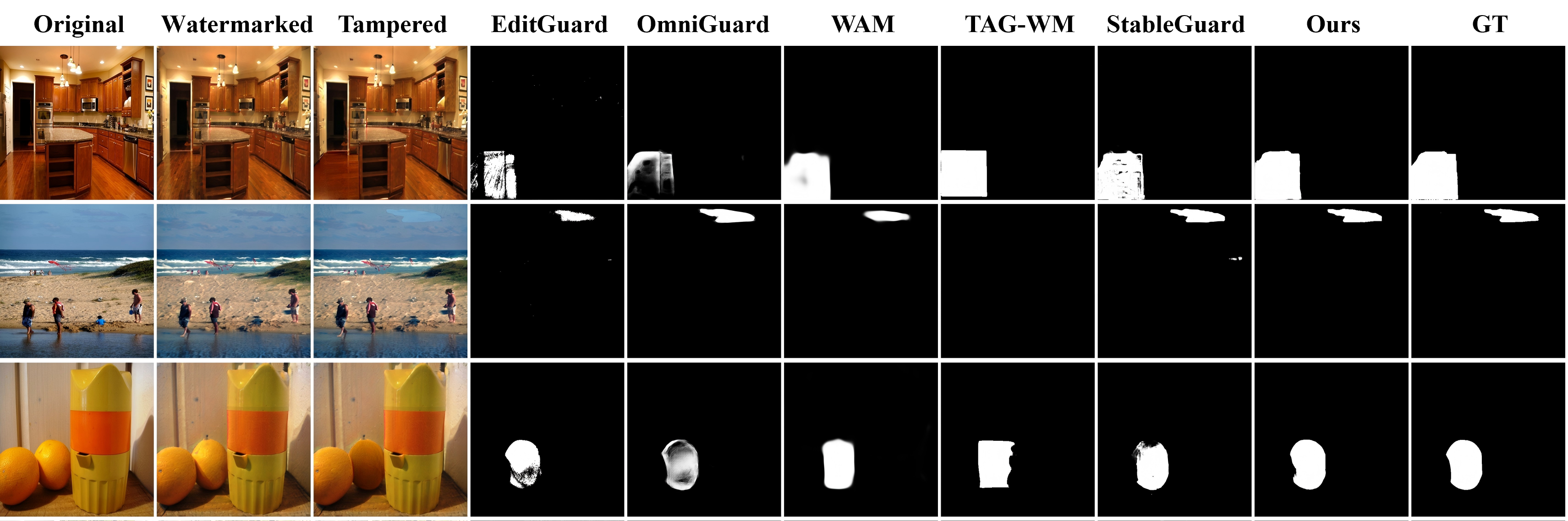}
  \caption{\textbf{Qualitative comparison of tamper localization on the generative-model-based tampering dataset~\cite{sd, sdxl, kandinsky, controlnet, lama}.} Our method generates precise localization masks that align closely with the ground truth (GT), effectively handling complex boundaries where baselines fail.}
  \label{fig:loc_img}
\end{figure*}

The overall and detailed comparisons are shown in Tab.~\ref{tab:combined_comparison} and Tab.~\ref{tab:aigc_dst}.
ROBIN++ significantly outperforms all competing methods in terms of localization accuracy, reaching a state-of-the-art average precision of 0.985. 
Passive detection approaches generally underperform methods that incorporate explicit, actively injected localization watermarks. 
Although StableGuard similarly targets the decoding stage, its joint optimization of robust and fragile objectives within the spatial domain forces a trade-off, compromising robustness (0.620 AUC) to achieve localization (0.978 precision).
In contrast, by injecting a spatial-domain content-aware perturbation that remains spectrally distinct from the robust signal, ROBIN++ ensures high sensitivity to semantic tampering while maintaining superior copyright robustness. This design effectively resolves the performance bottlenecks observed in existing dual-functional frameworks.

The quantitative superiority of ROBIN++ is further evidenced by the qualitative results in Fig.~\ref{fig:loc_img}. As illustrated, ROBIN++ generates precise localization masks that align closely with the ground truth (GT) across diverse scenes. 
While other baselines, such as StableGuard and OmniGuard, identify the general vicinity of tampering, they often produce fragmented results or suffer from false positives in non-tampered regions. Our bidirectional synergistic verification approach leverages the frequency-domain copyright watermark as a global reference to guide the localization process, further improving the precision of the generated masks.

\subsection{Image quality}
We further evaluate the perceptual quality of watermarked images using PickScore.
As reported in the last column of Tab.~\ref{tab:combined_comparison}, ROBIN++ achieves a PickScore of 19.124, which is comparable to existing state-of-the-art watermarking methods.
\new{Although it does not obtain the highest PickScore, the gap to the best-performing baseline is only 0.230 on the raw PickScore scale, corresponding to a relatively small preference-level shift under the PickScore softmax interpretation, as further analyzed in Appendix G. This slight reduction may arise because the additional watermark-related constraints introduced during generation mildly restrict the original sampling freedom, leading to small changes in preference-oriented visual factors. This pattern is consistently shared by StableGuard, which imposes watermark-related constraints on the latent-to-image decoding process. Overall, ROBIN++ preserves competitive visual quality while providing stronger dual-functional verification capability, including robust copyright authentication and fine-grained tamper localization. }

\subsection{Evaluation under combined degradation and tampering}
We further consider a more challenging scenario in which image degradations and malicious tampering coexist. This setting is inherently more difficult to evaluate, as benign distortions such as JPEG compression and additive noise can substantially interfere with accurate tamper localization, while the presence of localized tampering may also hinder reliable verification of robust copyright watermarks.

\begin{figure}[t]
  \centering
  \includegraphics[width=.9\linewidth]{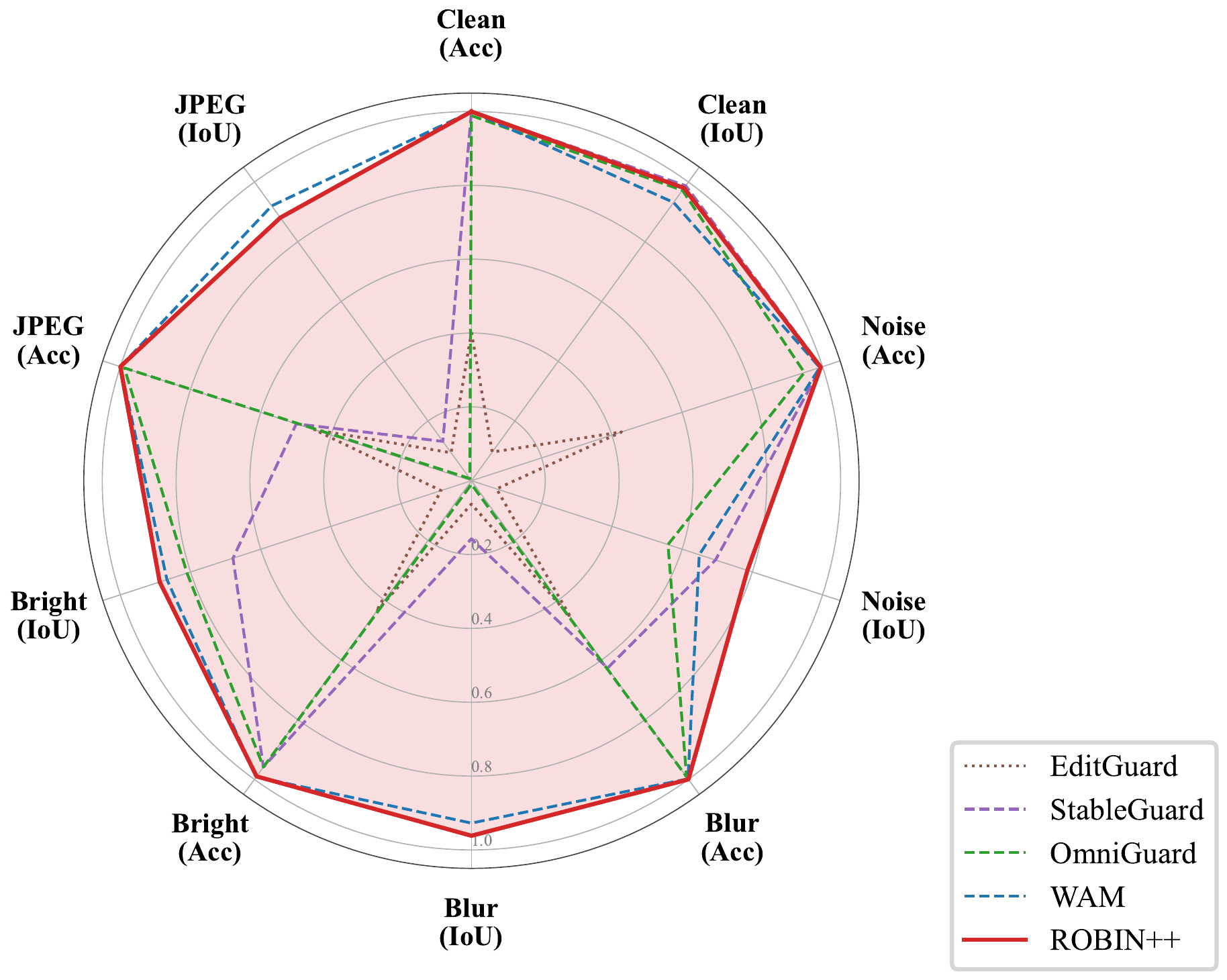}
  \caption{\textbf{Performance in complex scenarios with coexisting degradations and tampering.} ROBIN++ achieves the best trade-off between robustness and tamper localization accuracy compared to baselines, effectively distinguishing benign distortions from malicious attacks.}
  \label{fig:real_att}
\end{figure}

As shown in Fig.~\ref{fig:real_att}, our method achieves the best performance in both robustness and tamper localization under this combined setting. Specifically, ROBIN++ maintains an average localization IoU of 0.879 and a verification accuracy of 0.996 under combined degradation settings. This advantage stems from the synergistic interaction between the spatial- and frequency-domain watermarks. The robust frequency-domain watermark provides a global integrity cue that helps the localization module distinguish benign degradations from malicious manipulations. Conversely, accurate spatial tamper localization enables the robust watermark extractor to focus on reliable regions during verification. Moreover, under the same localization mask, frequency-domain watermarks are inherently easier to recover from partially intact regions than spatial-domain counterparts, further enhancing verification reliability \cite{stankovic2013image,cao2022robust}. Together, these properties allow our approach to maintain strong robustness and precise localization even in the presence of simultaneous degradations and tampering.

\subsection{Ablation study}
We conduct comprehensive ablation studies to analyze the contribution of each key component in ROBIN++.
All experiments are performed under the same evaluation protocol, and results are reported in Tab.~\ref{tab:ablation_resize}.

\subsubsection{Effect of Adversarial Optimization}
The first two rows validate the effectiveness of adversarial frequency-domain watermarking. Compared with directly embedding a frequency watermark, adversarial optimization improves the average robustness from 0.973 to 0.979 by jointly enhancing watermark strength and visual concealment.

\begin{table}[t]
  \centering
  \caption{Ablation Study Results. Note that ``Localization'' is measured by average IoU under five tampering types, while ``Robustness'' reports the average accuracy under clean, rotation, JPEG, and blur settings.}
  \renewcommand{\arraystretch}{1.3}
  \resizebox{1.0\linewidth}{!}{
    \begin{tabular}{ccccc|c|c|c}
      \hline
      \textbf{Freq} & \textbf{Adv} & \textbf{CAP} & \textbf{F2S} & \textbf{S2F} & \textbf{Localization} & \textbf{Robustness} & \textbf{Avg} \\
      \hline
      $\checkmark$ & & & & & 0.387 & \underline{0.973} & 0.680 \\
      $\checkmark$ & $\checkmark$ & & & & 0.543 & \textbf{0.979} & 0.761 \\
      $\checkmark$ & $\checkmark$ & $\checkmark$ & & & \underline{0.947} & 0.887 & 0.917 \\
      $\checkmark$ & $\checkmark$ & $\checkmark$ & $\checkmark$ & & \textbf{0.971} & 0.970 & \underline{0.971} \\
      $\checkmark$ & $\checkmark$ & $\checkmark$ & $\checkmark$ & $\checkmark$ & \textbf{0.971} & \underline{0.973} & \textbf{0.972} \\
      \hline
    \end{tabular}
  }
  \label{tab:ablation_resize}
\end{table}

\begin{table}[t]
\centering
\caption{Ablation study on key hyperparameters of the framework: CAP perturbation constraint ($\xi$) and frequency domain coverage ratio.}
\resizebox{\columnwidth}{!}{
\begin{tabular}{@{}llccc@{}}
\toprule
\multirow{2}{*}{\textbf{Param.}} & \multirow{2}{*}{\textbf{Value}} & \textbf{PickScore} & \textbf{Robustness} & \textbf{Local.} \\
& & ($\uparrow$) & (Acc. / AUC) $\uparrow$ & (Avg) $\uparrow$ \\
\midrule
\multirow{3}{*}{$\xi$} 
& 0.25 & 18.923 & 0.951 / 0.974 & \textbf{0.985} \\
& \textbf{0.50 (Def.)} & \textbf{19.124} & \textbf{0.973 / 0.984} & \textbf{0.985} \\
& 0.75 & 18.825 & 0.951 / 0.970 & 0.975 \\
\midrule
\multirow{4}{*}{\shortstack[l]{Freq.\\Cov.}} 
& 20\% & 19.034 & 0.950 / 0.968 & 0.976 \\
& 50\% & 18.777 & 0.961 / 0.978 & 0.984 \\
& \textbf{70\% (Def.)} & \textbf{19.124} & 0.973 / 0.984 & \textbf{0.985} \\
& 90\% & 18.599 & \textbf{0.977 / 0.992} & 0.981 \\
\bottomrule
\end{tabular}
}
\label{tab:hyperparameters}
\end{table}

\subsubsection{Effect of Localization Watermark}
The most significant localization improvement comes from the proposed Content-Aware Perturbation (CAP). Without CAP, the detector can only rely on passive statistical inconsistencies, leading to limited localization performance. After introducing the spatial-domain localization watermark, the localization IoU increases substantially from 0.543 to 0.947, demonstrating that active fragile watermarking provides explicit and learnable cues for precise tamper localization.

\subsubsection{Effect of Frequency-to-Spatial (F2S) Synergy}
When an additional frequency-domain branch is added, the localization 
IoU improves markedly from 0.947 to 0.971.
This result confirms that robust frequency-domain watermarks serve as an effective reference anchor, further enhancing the localization precision of the detector.
\new{The F2S synergy is particularly crucial under global degradations, where spatial traces are heavily smoothed. By leveraging the robust spectral signal as a global structural anchor, the detector can reliably distinguish uniform benign degradation from localized malicious manipulation.}

\subsubsection{Effect of Spatial-to-Frequency (S2F) Refinement}
S2F refinement consistently improves robustness from 0.970 to 0.973 by using the predicted tamper mask to suppress corrupted regions during copyright verification. These results show that the two watermarks are not merely decoded independently, but can mutually reinforce each other during verification.
\new{S2F refinement is most beneficial under large-scale local tampering. It acts as a latent-space quarantine mechanism, neutralizing corrupted regions to prevent adversarial spectral contamination from destroying global provenance verification.}

\subsubsection{Hyperparameter Analysis}
\new{To further validate our architectural choices, we analyze the framework's sensitivity to key hyperparameters (summarized in Tab.~\ref{tab:hyperparameters}). 
For the spatial branch, a perturbation constraint of $\xi=0.50$ provides the optimal balance. Stricter constraints ($\xi=0.25$) slightly reduce copyright robustness, while larger budgets ($\xi=0.75$) introduce unnecessary interference that degrades image quality.
For the frequency branch, the default $70\%$ coverage ratio optimally balances spectral redundancy and generation fidelity. Smaller ratios (e.g., $20\%$) lack the capacity needed for robust recovery and weaken the global prior required for spatial localization. Conversely, excessively large coverage ($90\%$) achieves higher robustness but overwrites semantically important frequency components, noticeably compromising perceptual quality.
}

\begin{figure}[t]
  \centering
  \includegraphics[width=.9\linewidth]{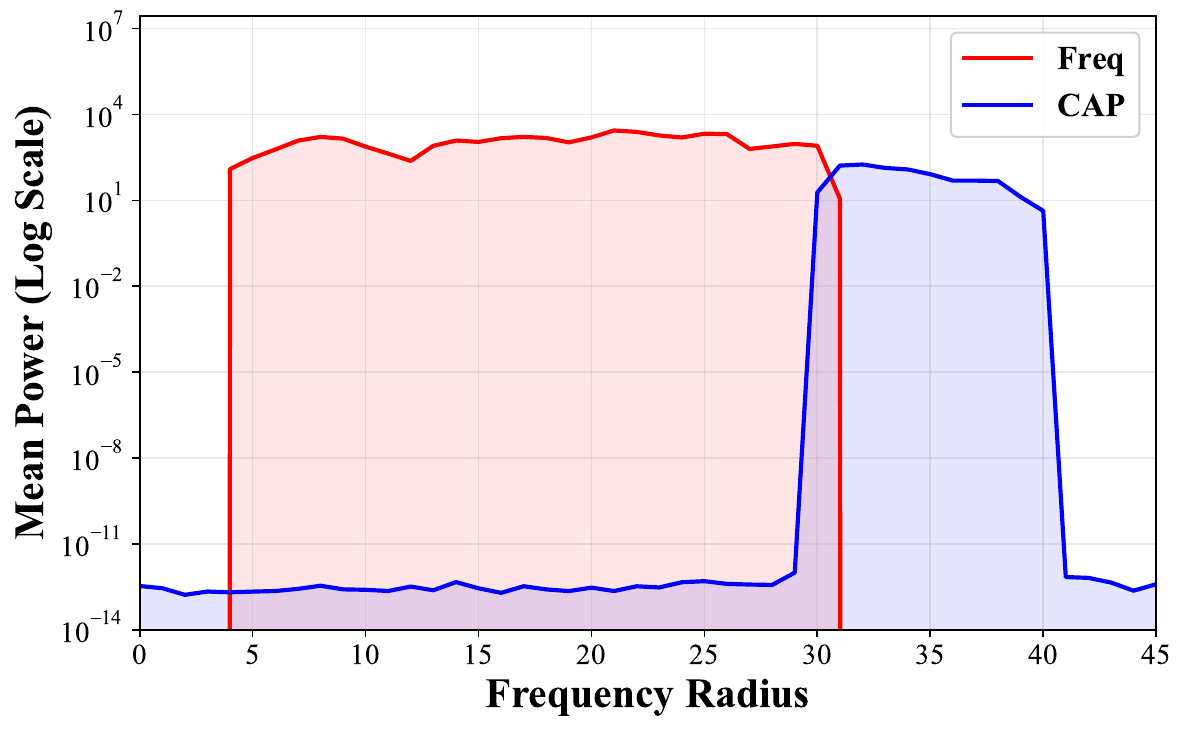}
  \caption{\textbf{Spectral separation of dual watermarks.} The robust copyright watermark (Freq) and fragile localization watermark (CAP) occupy distinct frequency bands, ensuring they function without mutual interference.}
  \label{fig:no_inter}
\end{figure}

\begin{figure}[t]
  \centering
  \includegraphics[width=1\linewidth]{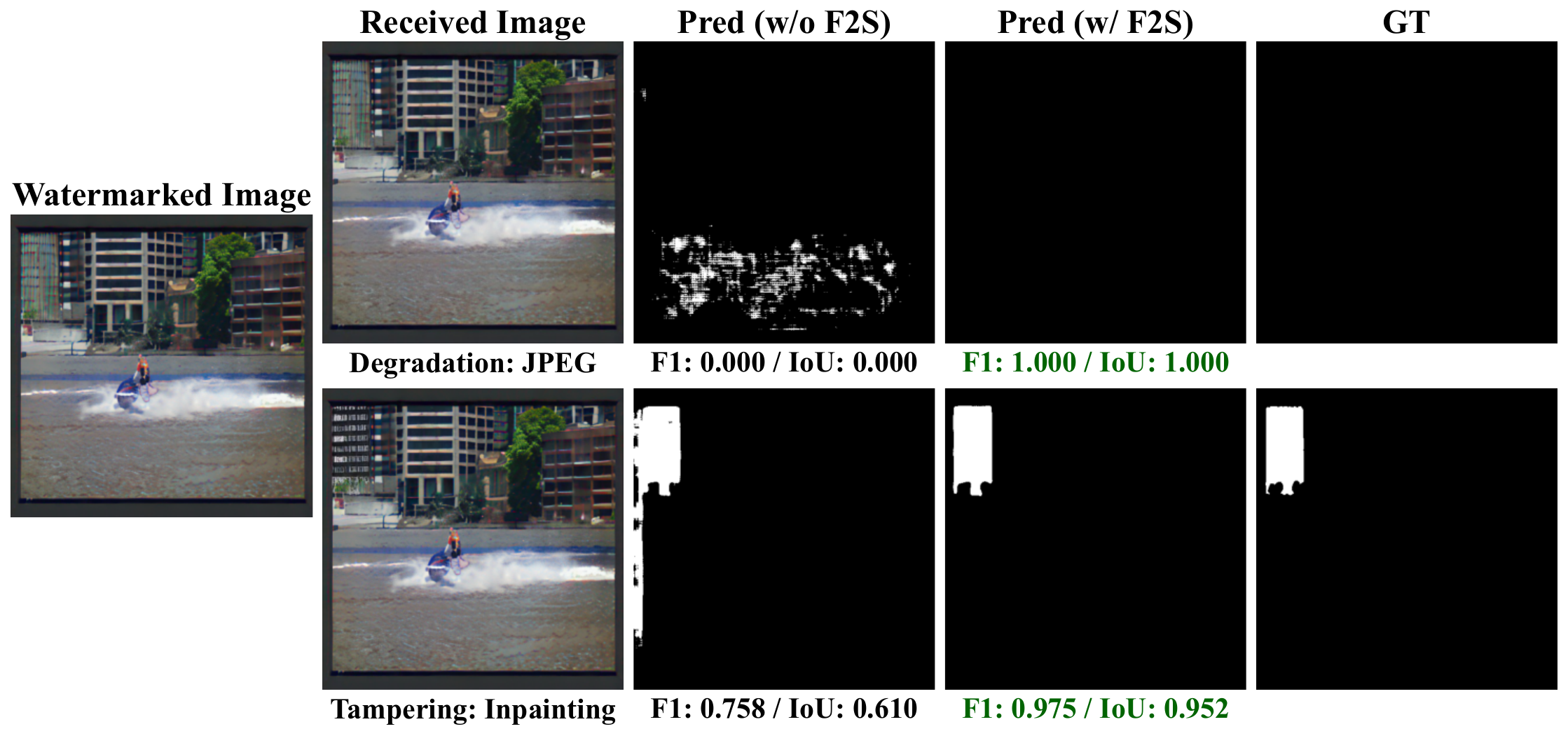}
  \caption{\textbf{Effectiveness of the frequency-to-spatial (F2S) synergy.} By leveraging frequency features, ROBIN++ achieves robust immunity to JPEG noise (top) and superior precision in inpainting localization (bottom).}
  \label{fig:freq2spac}
\end{figure}

\begin{figure}[t]
  \centering
  \includegraphics[width=1.0\linewidth]{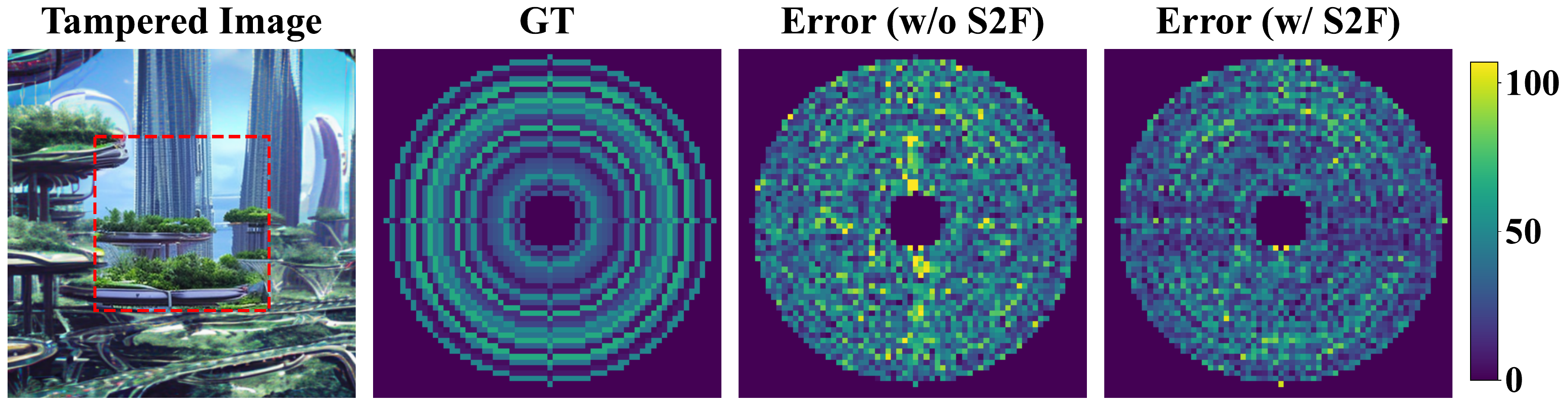}
  \caption{\textbf{Effectiveness of the spatial-to-frequency (S2F) refinement.} By utilizing the predicted tamper mask to suppress interference from the spliced region (red box), the extraction error is significantly reduced.}
  \label{fig:spac2freq}
\end{figure}

\begin{table}[t]
\centering
\caption{Quantitative comparison of zero-shot transferability trained on Stable Diffusion (SD) v2.1. 
}
\label{tab:migration_comparison}
\renewcommand{\arraystretch}{1.3}
\resizebox{\columnwidth}{!}{%
\begin{tabular}{c|ccc|ccc}
\hline
\multirow{2}{*}{\textbf{Method}} & \multicolumn{3}{c|}{\textbf{SD v2.1} $\to$ \textbf{SD v1.4}} & \multicolumn{3}{c}{\textbf{SD v2.1} $\to$ \textbf{SD v2.0}} \\ 
\cline{2-7}
 & F1 $\uparrow$ & AUC $\uparrow$ & IoU $\uparrow$ & F1 $\uparrow$ & AUC $\uparrow$ & IoU $\uparrow$ \\
\hline
StableGuard \cite{stableguard}& 0.007 & 0.492 & 0.004 & 0.971 & \textbf{0.999} & 0.953 \\
\textbf{ROBIN++} & \textbf{0.765} & \textbf{0.927} & \textbf{0.707} & \textbf{0.974} & \textbf{0.999} & \textbf{0.956} \\
\hline
\end{tabular}%
}
\end{table}

\subsection{Relationship between frequency and spatial watermark}

The frequency-domain and spatial-domain watermarks are designed to serve distinct functions without interfering with each other.
We analyze the distribution of these two watermarks in the frequency domain, as illustrated in Fig.~\ref{fig:no_inter}.
The copyright watermark (the red line) is embedded in the mid-to-low frequency range for robustness.
The tamper localization watermark (the blue line), as a sample-aware random perturbation, is primarily concentrated in the high-frequency region.
This separation ensures that the two watermarks do not interfere with each other, allowing for both robust copyright protection and precise tamper detection.

To verify the role of frequency-domain watermarking in assisting spatial localization, we evaluate the effectiveness of tamper localization under two scenarios: normal JPEG compression and malicious local tampering, as shown in Fig.~\ref{fig:freq2spac}.
The baseline without frequency-to-spatial (F2S) synergy struggles with compression artifacts, yielding 0 for both F1 and IoU, whereas our method achieves 1.000, indicating zero false positives. For inpainting attacks, our method significantly improves localization accuracy. These improvements validate that frequency-domain watermarks help the network differentiate between global operations and local manipulations.

We further evaluate the spatial-to-frequency (S2F) refinement strategy's impact on copyright watermark extraction accuracy (Fig.~\ref{fig:spac2freq}). With refinement, 
extraction error reduced significantly, as mean-based filling of tampered regions helps neutralize the distributional shift caused by the tampered content, making frequency-domain watermark extraction more robust and accurate.

\subsection{Transferability and Generalization of watermarks}
We further evaluate the zero-shot transferability of ROBIN++ by optimizing the watermarks on SD v2.1 and subsequently deploying it for watermark injection and image generation across different target models (SD v1.4, SD v2.0) without any adaptation~\cite{sd}. 
As shown in Tab.~\ref{tab:migration_comparison}, compared with similar in-generation watermarking methods, ROBIN++ demonstrates superior cross-model transferability. StableGuard fails on SD v1.4 (F1: 0.007) due to its reliance on residual adapters injected into the VAE decoder. 
In contrast, by injecting CAP into the shared latent space, our method exploits the structural consistency across diffusion versions. Consequently, we achieve a robust F1 of 0.765 on SD v1.4, confirming the advantage of operating on latent representations over modifying internal decoder parameters.

\new{To further examine the cross-architecture generalization of ROBIN++, we additionally train and evaluate it on Stable Diffusion XL (SDXL) and PixArt-$\alpha$.
As shown in Tab. XII and XIII in the Appendix, ROBIN++ achieves the best average robustness and inpainting localization while maintaining image quality within a reasonable range across both models.
This indicates that the proposed dual-domain watermarking design generalizes effectively to both large-scale U-Net-based diffusion models and transformer-based diffusion architectures.}

\subsection{Computational Efficiency}
\new{To evaluate the practical efficiency of ROBIN++, we quantify the computational overhead across different pipeline stages (a detailed breakdown is provided in Appendix Tab. XI). The offline model-specific preparation incurs a one-time cost of approximately 50.2 hours. During the online generation phase, our synergistic injection introduces a marginal relative overhead of merely +0.14 s/img and +1.0 GB VRAM compared to standard diffusion. The practical dual-branch verification is also highly efficient, requiring only ~0.78 s/img in total.}

\section{Conclusion, Limitations, and Future Work}

In this paper, we propose a Dual-Domain Synergistic watermarking framework for diffusion models, which simultaneously achieves robust copyright verification and precise tamper localization. By decoupling robustness and fragility at the injection stage and re-coupling them through synergistic verification, our method overcomes the inherent limitations of existing single-domain watermarking designs.
Specifically, we embed copyright watermarks in the frequency domain to leverage global stability, while introducing a spatial-domain localization watermark based on Content-Aware Perturbations (CAP) to preserve explicit spatial correspondence. Furthermore, we design a bidirectional synergistic verification strategy that integrates frequency-domain verification cues and spatial tamper predictions, allowing the two tasks to mutually reinforce each other.
Extensive experiments demonstrate that the proposed framework consistently outperforms existing methods in both watermark robustness and tamper localization accuracy. 
Overall, this work establishes a unified and scalable solution for dual-functional watermarking in diffusion-based generative models, and provides new insights into the co-design of robustness and sensitivity for secure and trustworthy generative media systems.

\noindent \textbf{Limitations and future work.}
Although the proposed framework demonstrates strong performance in unified verification of diffusion-generated images, it has certain limitations.
Specifically, our approach assumes access to the diffusion generation process, as both watermark signals are embedded during intermediate sampling.
This setting is well-suited for model owners and content providers who control the generation pipeline, but it may limit applicability in purely post-generation watermarking scenarios where such access is unavailable.
Beyond image generation, a promising future direction is to extend the proposed dual-domain synergistic paradigm to other generative modalities, such as video, 3D, and multi-view content, where temporal or geometric consistency offers new opportunities for robust and fine-grained watermarking.
We believe that the principle of decoupled injection with synergistic verification can serve as a general foundation for trustworthy generative systems.

\section*{Acknowledgment}

This work was partially supported by the National Natural Science Foundation of China under Grants 62372341 and 62225113, and the Innovative Research Group Project of Hubei Province under Grant 2024AFA017. It also received support from the New Cornerstone Science Foundation through the XPLORER PRIZE. Additional support was provided by WHU-Kingsoft Joint Lab.


\bibliography{robin++}
\bibliographystyle{IEEEtran}

\begin{IEEEbiography}[{\includegraphics[width=1in,height=1.25in,clip,keepaspectratio]{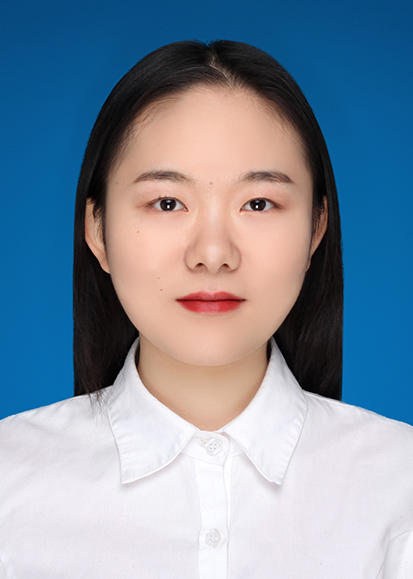}}]{Huayang Huang}
 received the master's degree from the School of Cyber Science and Engineering, Wuhan University (Wuhan, China), in 2024. She is currently working toward the PhD degree with the School of Computer Science, Wuhan University (Wuhan, China). Her research interests focus on safe generative AI.
\end{IEEEbiography}

\begin{IEEEbiography}[{\includegraphics[width=1in,height=1.25in,clip,keepaspectratio]{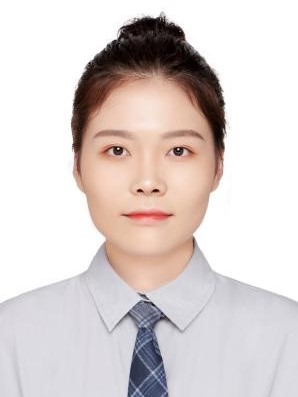}}]{Siqi Zeng} received a bachelor's degree from the School of Computer Science, Wuhan University (Wuhan, China) in 2025. She is currently pursuing a master’s degree in computer science at the School of Computer Science, Wuhan University (Wuhan, China). Her research interests mainly include safe generative AI.
\end{IEEEbiography}

\begin{IEEEbiography}[{\includegraphics[width=1in,height=1.25in, clip,keepaspectratio]{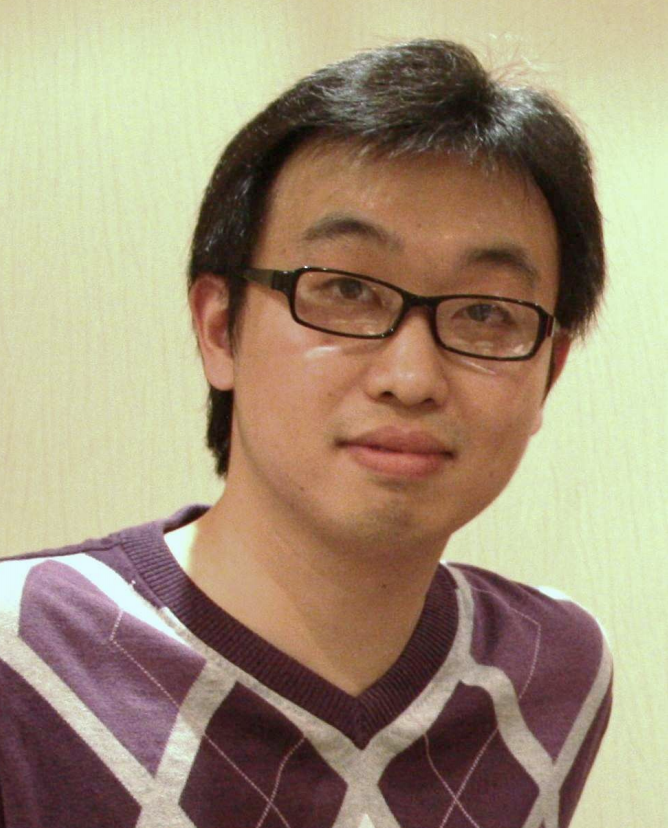}}]{Qian Wang} is a Professor in the School of Cyber Science and Engineering at Wuhan University, China. He was selected into the National High-level Young Talents Program of China, and listed among the World's Top 2\% Scientists by Stanford University. He also received the National Science Fund for Excellent Young Scholars of China in 2018. He has long been engaged in the research of cyberspace security, with focus on AI security, data outsourcing security and privacy, wireless systems security, and applied cryptography. He was a recipient of the 2018 IEEE TCSC Award for Excellence in Scalable Computing (early career researcher) and the 2016 IEEE ComSoc Asia-Pacific Outstanding Young Researcher Award. He has published 200+ papers, with 120+ publications in top-tier international conferences, including USENIX NSDI, ACM CCS, USENIX Security, NDSS, ACM MobiCom, ICML, etc., with 20000+ Google Scholar citations. He is also a co-recipient of 8 Best Paper and Best Student Paper Awards from prestigious conferences, including ICDCS, IEEE ICNP, etc. In 2021, his PhD student was selected under Huawei's  ``Top Minds'' Recruitment Program. He serves as Associate Editors for IEEE Transactions on Dependable and Secure Computing (TDSC) and IEEE Transactions on Information Forensics and Security (TIFS). He is a fellow of the IEEE, and a member of the ACM.
\end{IEEEbiography}

\begin{IEEEbiography}[{\includegraphics[width=1in,height=1.25in,clip,keepaspectratio]{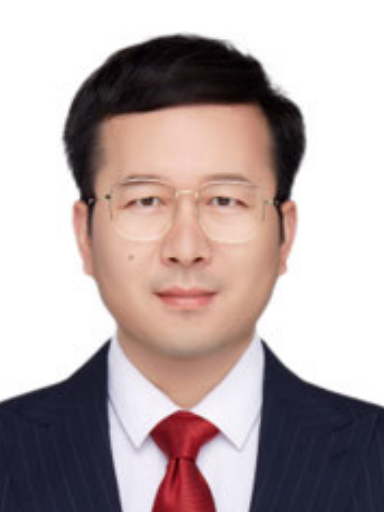}}]{Bo Du}
(Senior Member, IEEE) received the Ph.D. degree in photogrammetry and remote sensing from the State Key Laboratory of Information Engineering in Surveying, Mapping and Remote Sensing, Wuhan University, Wuhan, China, in 2010. He is currently a Professor with the National Engineering Research Center for Multimedia Software, Institute of Artificial Intelligence, Wuhan University, where he is also with the School of Computer Science. He has more than 60 research articles published in the IEEE TRANSACTIONS ON NEURAL NETWORKS AND LEARNING SYSTEMS, the IEEE TRANSACTIONS ON IMAGE PROCESSING, the IEEE TRANSACTIONS ON MULTIMEDIA, the IEEE TRANSACTIONS ON GEOSCIENCE AND REMOTE SENSING, the IEEE JOURNAL OF SELECTED TOPICS IN EARTH OBSERVATIONS AND APPLIED REMOTE SENSING, and the IEEE GEOSCIENCE AND REMOTE SENSING LETTERS. Thirteen of them are ESI hot articles or highly cited articles. His major research interests include pattern recognition, hyperspectral image processing, machine learning, and signal processing. He received the Highly Cited Researcher 2019 Award from the Web of Science Group, the Distinguished Paper Award from IJCAI 2018, the Best Paper Award of IEEE Whispers 2018, and the Champion Award of the IEEE Data Fusion Contest 2018. He received the Best Reviewer Award from IEEE GRSS for his service to the IEEE JOURNAL OF SELECTED TOPICS IN EARTH OBSERVATIONS AND APPLIED REMOTE SENSING in 2011 and the ACM Rising Star Award for his academic progress in 2015. He was the Session Chair of both International Geoscience and Remote Sensing Symposium (IGARSS) in 2016 and 2018 and the 4th IEEE GRSS Workshop on Hyperspectral Image and Signal Processing: Evolution in Remote Sensing (WHISPERS). He serves as an Area Chair for ICPR and IJCNN. He serves as an Associate Editor for Pattern Recognition and Neurocomputing and a Senior PC for IJCAI/AAAI/KDD. He also serves as a Reviewer for 20 science citation index (SCI) magazines, including IEEE TRANSACTIONS ON GEOSCIENCE AND REMOTE SENSING, IEEE TRANSACTIONS ON IMAGE PROCESSING, IEEE JOURNAL OF SELECTED TOPICS IN APPLIED EARTH OBSERVATIONS AND REMOTE SENSING, and IEEE GEOSCIENCE AND REMOTE SENSING LETTERS.
\end{IEEEbiography}

\begin{IEEEbiography}[{\includegraphics[width=1in,height=1.25in,clip,keepaspectratio]{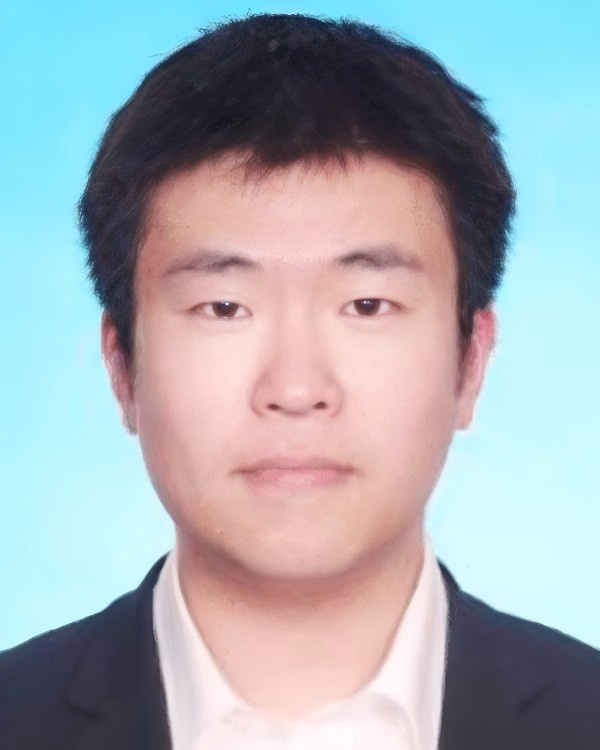}}]{Yu Wu}
 is a Professor with the School of Computer Science at Wuhan University, China. He received his Ph.D. degree from the University of Technology Sydney, Australia in 2021. From 2021 to 2022, he was a postdoc at Princeton University. His research interests are controllable generation and multi-modal perception. He was the recipient of AAAI New Faculty Highlight 2024 and Google PhD Fellowship 2020. He served as the Area Chair for CVPR, ICCV, ECCV, and NeurIPS, and also served as the Workshop Chair of CVPR 2023.
\end{IEEEbiography}

\newpage
\clearpage

{\appendices

\section{Localization on Non-Generative Local Edits}
While the main text focuses on generative-model-based inpainting, we further evaluate the tamper localization performance of ROBIN++ on non-generative local edits using the CASIA 2.0 dataset~\cite{dong2013casia}. We constructed a balanced subset comprising 500 copy-move images (where content is copied from within the same image) and 500 splicing images (where the inserted region comes from a different image). 
As shown in Tab.~\ref{tab:casia_balanced_tamper}, ROBIN++ consistently achieves the best performance across both traditional manipulation types. These results demonstrate that the spatial sensitivity of our Content-Aware Perturbation (CAP) extends robustly to realistic, Photoshop-style tampering.

\begin{table}[h]
    \centering
    \caption{Localization performance comparison on the balanced CASIA subset. The best and second-best results are highlighted in \textbf{bold} and \underline{underlined}, respectively.}
    \renewcommand{\arraystretch}{1.2}
    \resizebox{\columnwidth}{!}{
        \begin{tabular}{@{}l|ccc|ccc@{}}
            \hline
            \multirow{2}{*}{Method}
            & \multicolumn{3}{c|}{Copy-move}
            & \multicolumn{3}{c}{Splicing} \\
            \cline{2-7}
            & F1$\uparrow$ & AUC$\uparrow$ & IoU$\uparrow$
            & F1$\uparrow$ & AUC$\uparrow$ & IoU$\uparrow$ \\
            \hline
            EditGuard~\cite{editguard}
            & 0.847 & 0.990 & 0.757 
            & 0.871 & 0.989 & 0.791 \\
            OmniGuard~\cite{zhang2024omniguard} 
            & 0.786 & 0.967 & 0.722 
            & 0.672 & 0.949 & 0.595 \\
            WAM~\cite{wam}
            & 0.944 & \underline{0.999} & 0.903 
            & 0.955 & \underline{0.999} & 0.919 \\
            TAG-WM~\cite{tag}
            & 0.740 & 0.849 & 0.636 
            & 0.791 & 0.875 & 0.689 \\
            StableGuard~\cite{stableguard}
            & \underline{0.973} & \underline{0.999} & \underline{0.955} 
            & \underline{0.978} & \underline{0.999} & \underline{0.962} \\
            \hline
            ROBIN++
            & \textbf{0.988} & \textbf{1.000} & \textbf{0.977} 
            & \textbf{0.985} & \textbf{1.000} & \textbf{0.970} \\
            \hline
        \end{tabular}
    }
    \label{tab:casia_balanced_tamper}
\end{table}

\section{Quantitative Breakdown by Tamper Region Size.}
To provide a more informative view of localization performance, we further break down the evaluation metrics based on the size of the manipulated regions. We categorize the tamper masks into three distinct groups relative to the total image area: Small ($<$ 5\%), Medium (5\% - 15\%), and Large ($>$ 15\%).
As detailed in Tab.~\ref{tab:mask_size}, detecting minute tampering artifacts is inherently more difficult, leading to a noticeable performance drop in post-hoc methods like WAM on Small regions (IoU dropping to 0.794). In contrast, ROBIN++ demonstrates exceptional robustness across all scales. Notably, even though our framework employs a strategy that operates in the latent space using a downsampled pixel mask, it sustains superior precision on Small regions with an F1 score of 0.981 and an IoU of 0.964. This confirms that the Content-Aware Perturbation (CAP) injection provides highly sensitive, pixel-level cues that are not lost during latent-space translation.

\begin{table}[h]
\centering
\caption{
    Quantitative breakdown of tamper localization performance across different manipulated region sizes. 
    The best and second-best results are highlighted in \textbf{bold} and \underline{underlined}, respectively.
}
\resizebox{\linewidth}{!}{
\begin{tabular}{lcccccc}
\toprule
\multirow{2}{*}{\textbf{Method}} 
& \multicolumn{2}{c}{\textbf{Small ($<5\%$)}} 
& \multicolumn{2}{c}{\textbf{Medium ($5\% - 15\%$)}} 
& \multicolumn{2}{c}{\textbf{Large ($>15\%$)}} \\
\cmidrule(lr){2-3} \cmidrule(lr){4-5} \cmidrule(lr){6-7}
& \textbf{F1 $\uparrow$} & \textbf{IoU $\uparrow$} 
& \textbf{F1 $\uparrow$} & \textbf{IoU $\uparrow$} 
& \textbf{F1 $\uparrow$} & \textbf{IoU $\uparrow$} \\
\midrule
EditGuard~\cite{editguard} & 0.701 & 0.587 & 0.789 & 0.688 & 0.829 & 0.741 \\
OmniGuard~\cite{zhang2024omniguard} & 0.871 & 0.831 & 0.866 & 0.825 & 0.845 & 0.800 \\
WAM~\cite{wam} & 0.872 & 0.794 & 0.927 & 0.871 & 0.949 & 0.910 \\
TAG-WM~\cite{tag} & 0.528 & 0.402 & 0.778 & 0.666 & 0.873 & 0.784 \\
StableGuard~\cite{stableguard} 
& \underline{0.972} & \underline{0.946} 
& \underline{0.982} & \underline{0.965} 
& \underline{0.983} & \underline{0.967} \\
\textbf{ROBIN++ (Ours)} 
& \textbf{0.981} & \textbf{0.964} 
& \textbf{0.987} & \textbf{0.973} 
& \textbf{0.987} & \textbf{0.974} \\
\bottomrule
\end{tabular}
}
\label{tab:mask_size}
\end{table}

\section{Robustness against Advanced Attacks}
To supplement the evaluation of perceptible image transformations in the main text, we comprehensively evaluated ROBIN++ and state-of-the-art baselines under advanced security threats. The quantitative results are summarized in Tab.~\ref{tab:more_robust}.

\textbf{Imperceptible Attack.} We evaluate an imperceptible adversarial perturbation attack~\cite{an2024waves}, denoted as Adv. Noise. For each image, we optimize a bounded adversarial noise with an $\ell_\infty$ budget of $8/255$, such that the perturbed image is pushed away from the original image in the CLIP latent space. 
As shown in Tab.~\ref{tab:more_robust}, ROBIN++ achieves strong robustness to imperceptible adversarial noise.

\textbf{Frequency Attack.} We include three frequency-domain attacks that directly target the spectral space used by our watermark. Freq. Noise injects Gaussian noise (5\% of the image spectrum) into the frequency domain; Low Pass suppresses high-frequency information (with $9 \times 9$ kernel and standard deviation of 3.0); and Freq. Smooth smooths the spectral distribution to weaken structured frequency patterns (Gaussian blur with $\sigma$ = 1.5). 
The results show that ROBIN++ consistently achieves 1.000 under all three attacks, demonstrating that the optimized low-to-middle frequency watermark pattern is not easily destroyed by spectral perturbations or smoothing operations.

\textbf{Generative-AI-based Removal.} We evaluate recent reconstruction-based watermark removal attacks. Latent Remv.~\cite{muller2025black} uses SDXL to optimize the attacked image such that its $z_T$ representation deviates maximally from that of the original image. Diffusion Rec.~\cite{zhao2024invisible} reconstructs the image by re-noising and denoising it with SDXL for 60 steps. VAE Rec. applies a VAE encoding-decoding process to remove hidden signals. Rinse2x performs SDXL-based re-noising and denoising twice, representing a stronger iterative reconstruction attack. ROBIN++ also shows strong robustness under these challenging settings.

\textbf{Perturbation-removal.} Pert. Defeat~\cite{foerster2025lightshed} trains a universal perturbation-anomaly extractor to reconstruct and remove potential abnormal perturbations from the image. To rigorously test this, we fine-tuned the pre-trained checkpoints for 20 epochs on 2k clean-watermarked image pairs generated by each method (using different watermark messages from the test set). As shown in Tab.~\ref{tab:more_robust}, all evaluated watermarking methods remain highly robust against this attack. This is because Pert. Defeat is primarily designed to remove image-protection perturbations (e.g., Glaze-style adversarial cloaking). In contrast, robust watermarking methods are explicitly optimized to withstand small-range pixel-level perturbations, as evidenced by the strong robustness of all methods under imperceptible adversarial noise. Therefore, perturbation-level removal alone has limited effectiveness in destroying robust watermark signals.

\textbf{Black-box Forgery}. We evaluate a black-box latent forgery attack~\cite{muller2025black}, denoted as Latent Forg. 
Specifically, we use SDXL to optimize the latent $z_T$ of a random COCO image so that it becomes close to the $z_T$ representation of a given watermarked image, attempting to imitate the latent trace used for watermark verification without access to our watermarking model. We report the Forgery Success Rate (FSR), which measures the fraction of forged images that pass the watermark verification threshold; thus, lower values indicate stronger resistance to forgery.
As shown in Tab.~\ref{tab:more_robust}, ROBIN++ is the least vulnerable to the black-box forgery attack, outperforming all compared baselines. We attribute this to the in-process nature of ROBIN++: since the watermark is injected along the generation trajectory and tightly coupled with the original image semantics and denoising dynamics, directly forcing an unrelated cover image toward the target latent watermark trace is difficult to achieve reliably without substantially disturbing the cover image content.

\section{Degradation Profiling}
To clarify our operational limits, ``catastrophic degradation" in our framework represents a verification failure where the extracted watermark difference exceeds the empirical authentication threshold (set to 0.5).
To map this failure mode, we evaluate a subset of 500 images under progressive Gaussian noise addition. As shown in Fig.~\ref{fig:degradation_profile}, ROBIN++ demonstrates a highly resilient degradation profile. While most baseline methods exhibit a continuous decline in verification accuracy starting from mild noise levels, ROBIN++ sustains a robust accuracy plateau (Bit Acc $>$ 0.97) until the PSNR drops to approximately 15.0. Beyond this threshold of severe distortion, the verification accuracy declines progressively, yet our framework consistently outperforms all compared baselines across the entire evaluated noise spectrum.

\begin{figure}[h]
    \centering
    \includegraphics[width=.9\linewidth]{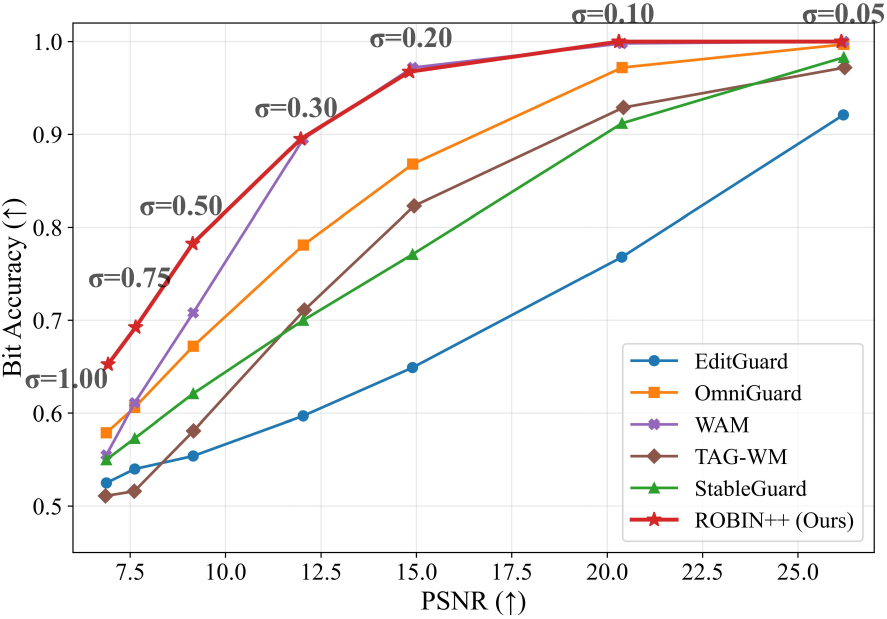}
    \caption{Degradation profile of ROBIN++ under progressive Gaussian noise with varying intensities ($\sigma \in \{0.05, 0.10, 0.20, 0.30, 0.50, 0.75, 1.00\}$).
    }
    \label{fig:degradation_profile}
\end{figure}

\section{Qualitative Analysis of Challenging and Failure Cases}

To provide a balanced view of our model's behavior, we visualize difference maps and error maps under highly challenging scenarios where the framework reaches its operational limits.

\textbf{Failure Mode 1: Severe Degradation Overriding Fragility (Error Maps)}: While our synergistic detector accurately localizes tampering under standard conditions, it struggles when local tampering is compounded by catastrophic global degradation. Fig.~\ref{fig:error_map} shows an error map where the watermarked image was subjected to a malicious inpainting attack followed by extreme JPEG compression or blurring. In the error map, correctly predicted pixels are white, False Positives are red, and False Negatives are blue. Because the extreme distortion severely smooths out the fragile watermark ($w_{\text{loc}}$), the spatial detector loses its primary cue. Consequently, the error map exhibits significant False Negatives (blue regions) inside the tampered area, leading to incorrect mask prediction.

\textbf{Failure Mode 2: Extreme Flat Regions (Difference Maps)}: Fig.~\ref{fig:residual} visualizes the $10\times$ amplified pixel difference between the clean and watermarked images to expose the CAP injection pattern. Our CAP generator is designed to hide perturbations within local textures. However, in challenging cases where the image features vast, completely flat, and textureless regions (e.g., a perfectly uniform background), the lack of structural masking forces the perturbation to become slightly more pronounced. 

\begin{figure}[h]
    \centering
    \includegraphics[width=\linewidth]{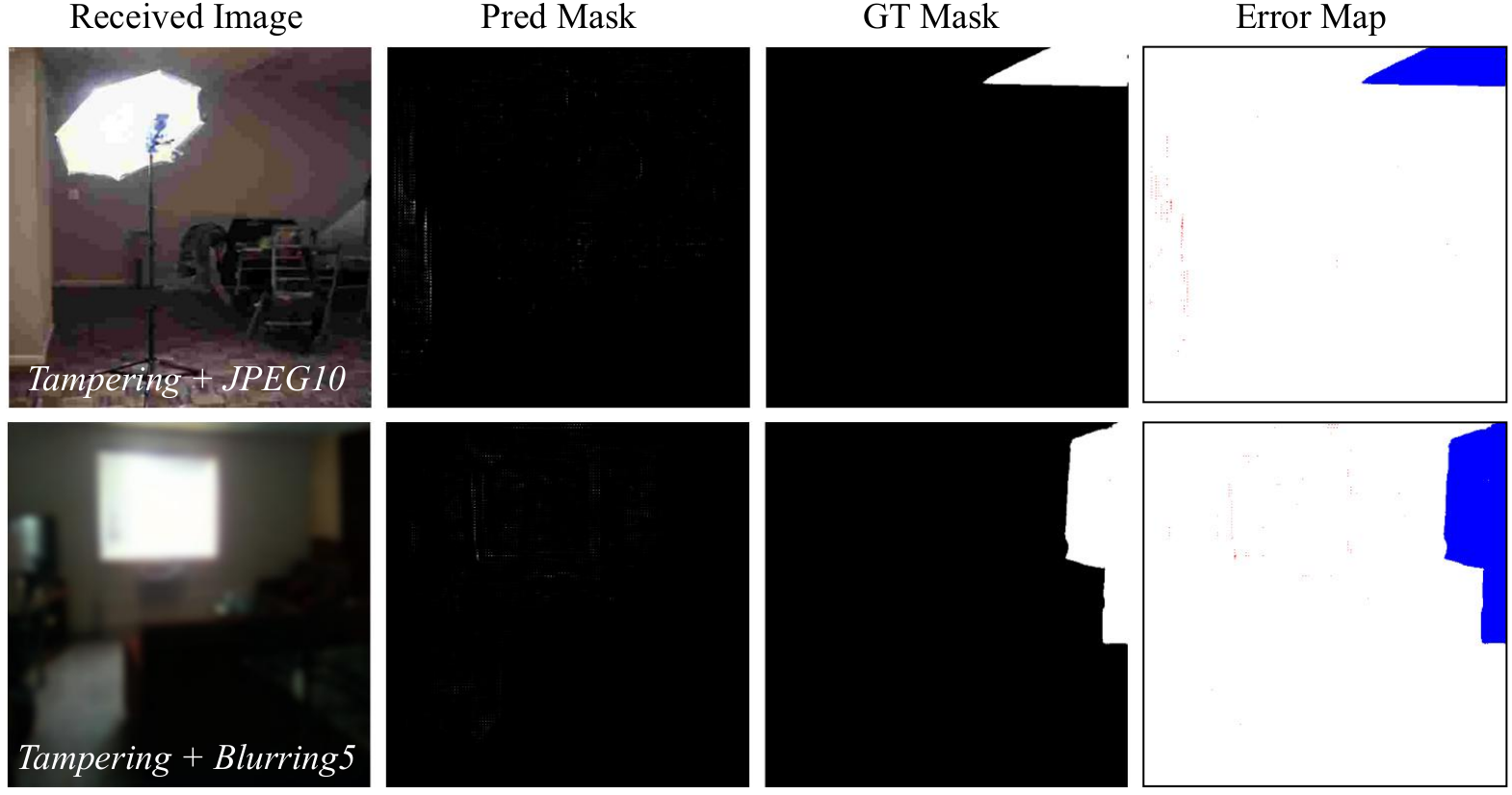}
    \caption{Error maps for tamper localization under challenging degradation conditions. False positives are marked in red, and false negatives are marked in blue.}
    \label{fig:error_map}
\end{figure}

\begin{figure}[h]
    \centering
    \includegraphics[width=.8\linewidth]{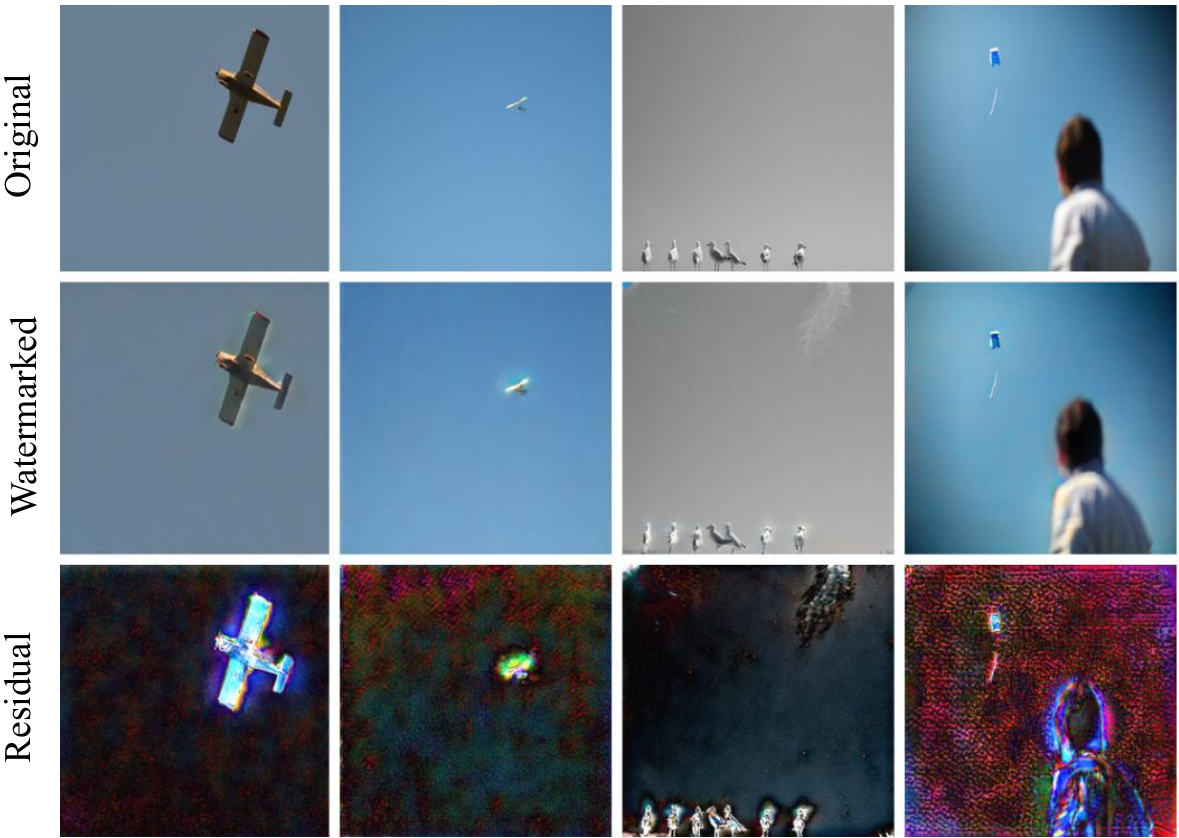}
    \caption{Amplified residual maps between original and watermarked images in challenging cases.}
    \label{fig:residual}
\end{figure}

\section{Adaptive Injection Scheduling}
\label{app:adaptive_timestep}
We further implement an adaptive watermark injection scheduler during offline optimization. Specifically, we maintain a historical score $r(t)$ for each candidate timestep in $\mathcal{T}$, where the score is defined as the negative offline optimization loss. The timestep scores are updated by exponential moving average: $
q_t \leftarrow (1-\rho)q_t+\rho r(t),
$
where $q_t$ is the historical score for timestep $t$, and $\rho$ controls the update rate. The sampling distribution is then recomputed using a softmax function with $\epsilon$-greedy exploration:
$
\pi(t)=
(1-\epsilon)
\frac{\exp(q_t/\tau)}
{\sum_{t'\in\mathcal{T}}\exp(q_{t'}/\tau)}
+
\epsilon\frac{1}{|\mathcal{T}|}.
$ In this way, timesteps that better balance watermark strength and visual fidelity are sampled more frequently during optimization, while sufficient exploration is retained.
As shown in Tab.~\ref{tab:adaptive}, the adaptive scheduler achieves performance close to the fixed injection setting, and changing the training configurations only leads to minor performance variations. We further observe that the learned sampling distribution concentrates on the intermediate region around $0.2T$--$0.3T$, which is consistent with our default injection window. Therefore, we keep the fixed window as the default setting for simplicity and reproducibility.

\begin{table}[h]
\centering
\caption{Comparison between fixed injection scheduling and the proposed Adaptive Injection Scheduling. Peak $t_\text{inj}$ denotes the timestep with the highest final sampling probability for the adaptive scheduler.}
\label{tab:adaptive}
\resizebox{\linewidth}{!}{
\setlength{\tabcolsep}{2pt}
\begin{tabular}{lcccc}
\toprule
\multirow{2}{*}{\textbf{Injection Scheduling}} & \textbf{Peak}
& \textbf{Robustness}
& \textbf{Localization}
& \multirow{2}{*}{\textbf{PickScore}$\uparrow$} \\
&$\boldsymbol{t_\text{inj}}$& \textbf{(Bit Acc. / AUC)} $\uparrow$ & \textbf{((F1+AUC+IoU)/3)} $\uparrow$ & \\
\midrule
Fixed (50 samples, cfg=7.5) & 300
& \textbf{0.973} / \underline{0.984} & \underline{0.985} & 19.124 \\
Adaptive (50 samples, cfg=7.5) & 280
& \underline{0.966} / \textbf{0.985} & \textbf{0.987} & 19.142 \\
Adaptive (200 samples, cfg=7.5) & 220
& 0.957 / 0.975 & \textbf{0.987} & \textbf{19.154} \\
Adaptive (50 samples, cfg=5.5) & 240
& 0.961 / 0.978 & \textbf{0.987} & \underline{19.151} \\
Adaptive (other 50 samples, cfg=7.5) & 280
& \underline{0.966} / \textbf{0.985} & \textbf{0.987} & 19.142 \\
\bottomrule
\end{tabular}
}
\end{table}

\section{PickScore Scale Analysis}
\label{app:pickscore}

To better interpret the PickScore results in Tab.~\ref{tab:combined_comparison}, we provide an additional analysis of the PickScore scale. PickScore is a CLIP-based preference model trained with a pairwise preference objective. For two images evaluated under the same prompt, the raw PickScore outputs can be interpreted as logits and converted into a model-implied pairwise preference probability through a softmax function:
\begin{equation}
P(i \succ j)=\frac{\exp(s_i)}{\exp(s_i)+\exp(s_j)}
=\frac{1}{1+\exp(-(s_i-s_j))},
\end{equation}
where $s_i$ and $s_j$ denote the raw PickScore values of two images. In Tab.~\ref{tab:combined_comparison}, the best PickScore among the compared methods is 19.354, while ROBIN++ obtains 19.124. The raw score gap is therefore $\Delta=19.354-19.124=0.230$, which corresponds to an approximate model-implied preference probability of 55.7\% for the higher-scored baseline in a hypothetical paired comparison. This suggests that the numerical gap reflects a relatively small preference-level difference under the PickScore model.

We further conduct a lightweight calibration experiment on non-watermarked generated images to examine how PickScore responds to mild, semantics-preserving appearance changes. Specifically, PickScore changes by $-0.155$ under JPEG compression with quality 90, by $-0.107$ under Gaussian noise with $\sigma=5/255$, and by $+0.010$ under a mild unsharp mask. These results indicate that PickScore is sensitive to subtle low-level appearance variations.
Therefore, the approximately 1\% PickScore gap of ROBIN++ should be interpreted as a small preference-level shift under a sensitive CLIP-based preference metric.

\begin{table}[h]
\centering
\caption{Summary of Key Notations}
\renewcommand{\arraystretch}{0.85}
\resizebox{\linewidth}{!}{
\begin{tabular}{@{}lp{0.75\linewidth}@{}}
\toprule
\textbf{Notation} & \textbf{Description} \\ \midrule
$z_t$, $z_0^*$, $z_w$, $\tilde{z}_0$ & Noisy, copyright-embedded, dual-watermarked, and rectified latent \\
$w_{\text{id}}$, $\hat{w}_{\text{id}}$ & Embedded robust copyright watermark and its extracted estimate \\
$w_{\text{loc}}$, $w_p$ & Fragile localization watermark, and learnable prompt guidance \\
$\mathcal{I}(\cdot)$, $\mathcal{T}(\cdot)$ & Dual-domain injection function, and compound degradation process \\
$\mathcal{G}_{\text{loc}}(\cdot)$, $\mathcal{D}_{\text{loc}}(\cdot)$ & Content-aware perturbation generator, and synergistic detector \\
$Dec(\cdot)$ & VAE decoder mapping latent representations to pixel space \\
$t_{\text{inj}}$, $\Omega$ & Intermediate injection timestep window, and frequency-domain mask \\
$M$, $\hat{M}$, $\hat{M}_z$ & GT tampering mask, its prediction, and downsampled latent mask \\
$\hat{P}$, $\tau_{\text{loc}}$ & Predicted probability map, and localization binarization threshold \\ \bottomrule
\end{tabular}
}
\label{tab:notation}
\end{table}
\begin{table*}[t]
\centering
\caption{
    Comprehensive evaluation of the computational overhead and practical efficiency of the ROBIN++ framework. 
    All measurements were conducted on a single NVIDIA RTX 4090 (48GB) GPU using a batch size of 1.
}
\resizebox{\textwidth}{!}{
\begin{tabular}{llccc}
\toprule
\textbf{Pipeline Stage} & \textbf{Specific Component} & \textbf{Time Cost / Runtime} & \textbf{Memory Overhead (VRAM)} & \textbf{Frequency of Cost} \\
\midrule

\multirow{2}{*}{\shortstack[l]{Offline Preparation \\ \textit{(Model-specific training)}}} 
& Adversarial Optimization & $\sim$0.20 Hours & 12.7 GB & One-time \\
& CAP Generator Training & $\sim$50.00 Hours & 41.7 GB & One-time \\
\midrule
\multirow{3}{*}{\shortstack[l]{Online Injection \\ \textit{(Generation pipeline)}}} 
& Standard Diffusion (Baseline) & $\sim$1.64 s / img & 3.9 GB & Per generation \\
& ROBIN++ Synergistic Injection & $\sim$1.78 s / img & 4.9 GB & Per generation \\
\cmidrule{2-5}
& \textit{Relative Overhead $\Delta$} & \textit{+0.14 s / img} & \textit{+1.0 GB} & - \\
\midrule
\multirow{3}{*}{\shortstack[l]{Practical Verification \\ \textit{(Dual-branch detection)}}} 
& Copyright Extraction (Freq.) & $\sim$0.30 s / img & 3.8 GB & Per verification \\
& Tamper Localization (Spatial) & $\sim$0.48 s / img & 19.9 GB & Per verification \\
\cmidrule{2-5}
& \textit{Total Verification Cost} & \textit{$\sim$0.78 s / img} & \textit{19.9 GB} & - \\
\bottomrule
\end{tabular}
}
\label{tab:overhead}
\end{table*}

\begin{table*}[t]
    \centering
    \caption{Evaluation of ROBIN++ Generalization to Stable Diffusion XL (SDXL). 
    We include the baselines with reliable implementations on SDXL.
    The best and second-best results are highlighted in \textbf{bold} and \underline{underlined}, respectively.}
    \renewcommand{\arraystretch}{1.2}
    \resizebox{\linewidth}{!}{
        \begin{tabular}{@{}l|c|ccccc|cccc|c@{}}
            \hline
            \multirow{2}{*}{\textbf{Method}}& \multirow{2}{*}{\textbf{Ref.}}
            & \multicolumn{5}{c|}{\textbf{Robustness (Bit Acc~$\uparrow$ / AUC~$\uparrow$)}} 
            & \multicolumn{4}{c|}
            {\makebox[0.32\linewidth][c]{\textbf{Inpainting Localization} \textbf{((F1+AUC+IoU)/3) $\uparrow$}}}
            & \textbf{Quality} \\ 
          
            \cline{3-12}
            
            & & Clean 
            & Rotation
            & JPEG 
            & Blur 
            & \textbf{Avg} 
            & \makebox[0.08\linewidth][c]{SDXL}
            & \makebox[0.08\linewidth][c]{Kandinsky}
            & \makebox[0.08\linewidth][c]{LaMa}
            & \makebox[0.08\linewidth][c]{\textbf{Avg}} 
            & PickScore \\ 
            \hline
            
            EditGuard & CVPR'24
            & \underline{0.834} / \underline{0.983}
            & \underline{0.505} / \underline{0.553}
            & 0.498 / \underline{0.512}
            & 0.455 / 0.286
            & 0.573 / 0.584
            & \makebox[0.08\linewidth][c]{0.837}
            & \makebox[0.08\linewidth][c]{0.830}
            & \makebox[0.08\linewidth][c]{0.841}
            & \makebox[0.08\linewidth][c]{0.836}
            & \underline{21.212} \\

            OmniGuard & CVPR'25
            & \textbf{1.000} / \textbf{1.000}
            & 0.493 / 0.478
            & \textbf{1.000} / \textbf{1.000}
            & \textbf{1.000} / \textbf{1.000}
            & \underline{0.873} / \underline{0.870}
            & \makebox[0.08\linewidth][c]{0.925}
            & \makebox[0.08\linewidth][c]{0.942}
            & \makebox[0.08\linewidth][c]{0.683}
            & \makebox[0.08\linewidth][c]{0.850}
            & \textbf{21.246} \\

            WAM & ICLR'25
            & \textbf{1.000} / \textbf{1.000}
            & 0.479 / 0.405
            & \underline{0.990} / \textbf{1.000}
            & \textbf{1.000} / \textbf{1.000}
            & 0.867 / 0.851
            & \makebox[0.08\linewidth][c]{0.950}
            & \makebox[0.08\linewidth][c]{0.946}
            & \makebox[0.08\linewidth][c]{0.953}
            & \makebox[0.08\linewidth][c]{0.950}
            & 21.028 \\

            StableGuard & NeurIPS'25
            & \textbf{1.000} / \textbf{1.000}
            & 0.498 / 0.478
            & 0.499 / 0.480
            & \underline{0.500} / \underline{0.490}
            & 0.624 / 0.612
            & \makebox[0.08\linewidth][c]{\underline{0.966}}
            & \makebox[0.08\linewidth][c]{\underline{0.965}}
            & \makebox[0.08\linewidth][c]{\underline{0.967}}
            & \makebox[0.08\linewidth][c]{\underline{0.966}}
            & 20.972 \\

            \hline

            ROBIN++ & -
            & \textbf{1.000} / \textbf{1.000}
            & \textbf{0.673} / \textbf{0.650}
            & \textbf{1.000} / \textbf{1.000}
            & \textbf{1.000} / \textbf{1.000}
            & \textbf{0.918} / \textbf{0.913}
            & \makebox[0.08\linewidth][c]{\textbf{0.970}}
            & \makebox[0.08\linewidth][c]{\textbf{0.968}}
            & \makebox[0.08\linewidth][c]{\textbf{0.972}}
            & \makebox[0.08\linewidth][c]{\textbf{0.970}}
            & 21.009 \\
            \hline
      
        \end{tabular}
    }
    \label{tab:sdxl_comparison}
\end{table*}

\begin{table*}[t]
    \centering
    \caption{Evaluation of ROBIN++ Generalization to PixArt-$\alpha$. We include the baselines with reliable implementations on PixArt-$\alpha$. The best and second-best results are highlighted in \textbf{bold} and \underline{underlined}, respectively.}
    \renewcommand{\arraystretch}{1.2}
        \resizebox{\linewidth}{!}{
        \begin{tabular}{@{}l|c|ccccc|cccc|c@{}}
            \hline
            \multirow{2}{*}{\textbf{Method}} & \multirow{2}{*}{\textbf{Ref.}}
            & \multicolumn{5}{c|}{\textbf{Robustness (Bit Acc~$\uparrow$ / AUC~$\uparrow$)}} 
            & \multicolumn{4}{c|}
            {\makebox[0.32\linewidth][c]{\textbf{Inpainting Localization} \textbf{((F1+AUC+IoU)/3) $\uparrow$}}}
            & \textbf{Quality} \\ 

            \cline{3-12}

            & & Clean 
            & Rotation
            & JPEG 
            & Blur 
            & \textbf{Avg} 
            & \makebox[0.08\linewidth][c]{SDXL}
            & \makebox[0.08\linewidth][c]{Kandinsky}
            & \makebox[0.08\linewidth][c]{LaMa}
            & \makebox[0.08\linewidth][c]{\textbf{Avg}} 
            & PickScore \\ 
            \hline

            EditGuard & CVPR'24
            & \textbf{1.000} / \textbf{1.000} & 0.487 / 0.456 & 0.701 / 0.964 & 0.472 / 0.414 & 0.665 / 0.709
            & \makebox[0.08\linewidth][c]{0.666} & \makebox[0.08\linewidth][c]{0.606} & \makebox[0.08\linewidth][c]{0.662}
            & \makebox[0.08\linewidth][c]{0.645}
            & \textbf{20.588} \\

            OmniGuard & CVPR'25
            & \textbf{1.000} / \textbf{1.000} & \underline{0.502} / 0.497 & \textbf{0.994} / \textbf{1.000} & \underline{0.997} / \textbf{1.000} & \underline{0.873} / 0.874
            & \makebox[0.08\linewidth][c]{\underline{0.975}} & \makebox[0.08\linewidth][c]{\textbf{0.963}} & \makebox[0.08\linewidth][c]{0.885}
            & \makebox[0.08\linewidth][c]{0.941}
            & \underline{20.558} \\

            WAM & ICLR'25
            & \textbf{1.000} / \textbf{1.000} & 0.496 / \underline{0.510} & 0.944 / \underline{0.998} & \textbf{1.000} / \textbf{1.000} & 0.860 / \underline{0.877}
            & \makebox[0.08\linewidth][c]{0.955} & \makebox[0.08\linewidth][c]{0.956} & \makebox[0.08\linewidth][c]{0.955}
            & \makebox[0.08\linewidth][c]{0.955}
            & 20.455 \\

            StableGuard & NeurIPS'25
            & \textbf{1.000} / \textbf{1.000} & 0.498 / 0.486 & 0.495 / 0.491 & 0.490 / 0.458 & 0.621 / 0.609
            & \makebox[0.08\linewidth][c]{0.970} & \makebox[0.08\linewidth][c]{\underline{0.961}} & \makebox[0.08\linewidth][c]{\underline{0.958}}
            & \makebox[0.08\linewidth][c]{\underline{0.963}}
            & 20.431 \\

            \hline

            ROBIN++ & -
            & \textbf{1.000} / \textbf{1.000} & \textbf{0.908} / \textbf{0.944} & \underline{0.980} / 0.986 & 0.960 / \underline{0.990} & \textbf{0.962} / \textbf{0.980}
            & \makebox[0.08\linewidth][c]{\textbf{0.977}} & \makebox[0.08\linewidth][c]{\underline{0.961}} & \makebox[0.08\linewidth][c]{\textbf{0.959}}
            & \makebox[0.08\linewidth][c]{\textbf{0.966}}
            & 20.472 \\
            \hline

        \end{tabular}
    }
    \label{tab:dit_comparison}
\end{table*}

}

%

\vfill

\end{document}